\documentclass{article}

\usepackage{microtype}
\usepackage{graphicx}
\usepackage{subcaption}
\usepackage{booktabs}
\usepackage{makecell}
\usepackage{enumitem}
\usepackage[utf8]{inputenc}
\usepackage{textgreek}
\usepackage{tcolorbox}
\usepackage{placeins}
\usepackage{caption}
\usepackage{float}

\usepackage{hyperref}

\usepackage[accepted]{icml2026}

\usepackage{amsmath}
\usepackage{amssymb}
\usepackage{mathtools}
\usepackage{amsthm}

\usepackage[capitalize,noabbrev]{cleveref}

\theoremstyle{plain}

\theoremstyle{definition}

\theoremstyle{remark}

\usepackage[disable,textsize=tiny]{todonotes}

\makeatletter
\renewcommand{\ICML@appearing}{%
  43rd International Conference on Machine Learning 2026, Seoul, South Korea, Workshop on Forecasting as a New Frontier of Intelligence.%
}
\makeatother

\begin{document}

\twocolumn[
  \icmltitle{Quantizing Time-Series Models As Dynamical Systems: Trajectory-Based Quantization Sensitivity Score}



  \icmlsetsymbol{equal}{*}

  \begin{icmlauthorlist}
    \icmlauthor{Mariya Pavlova}{yyy}
    \icmlauthor{Harrison Bo Hua Zhu}{sch}
    \icmlauthor{Lidia Vitanova}{lid}
    \icmlauthor{Elizaveta Semenova}{abv}
    \icmlauthor{Yingzhen Li}{yyy}
  \end{icmlauthorlist}

  \icmlaffiliation{yyy}{Department of Computing, Imperial College London, London, United Kingdom}
  \icmlaffiliation{sch}{Department of Public Health, University of Copenhagen, Copenhagen, Denmark}
 \icmlaffiliation{abv}{School of Public Health, Imperial College London, London, United Kingdom}
 \icmlaffiliation{lid}{GATE Institute, Sofia University St. Kliment Ohridski, Sofia, Bulgaria}
  \icmlcorrespondingauthor{Mariya Pavlova}{m.pavlova22@imperial.ac.uk}
  \icmlkeywords{Machine Learning, Quantization, Time-Series Models, Dynamical Models, Foundation Models}

  \vskip 0.3in
]



\printAffiliationsAndNotice{}  

\begin{abstract}
We introduce the Trajectory-based Quantization Sensitivity Score (TQS), a metric that reframes post-training quantization (PTQ) through the lens of dynamical-systems stability. By modeling the network’s rollout as a discrete-time dynamical system, TQS characterizes how quantization-induced errors propagate and amplify over the rollout horizon. Unlike conventional PTQ methods, where sensitivity analysis is often coupled to the quantization procedure, TQS enables a priori sensitivity estimation decoupled from quantizer selection  and bit-width assignment. This separation allows for quantization budget planning even for black-box or compiled networks with fused operators. Building on this, we present TQS-PTQ, a flexible mixed-precision framework that requires no calibration data or costly second-order approximations. Our experiments show that a dynamical-systems perspective provides a robust, high-performing pathway for low-precision deployment in resource-constrained settings.

\end{abstract}

\section{Introduction}
\label{sec:introduction}
Post-training quantization (PTQ) replaces full-precision weights, and in some settings activations, with low-precision representations after training, reducing memory footprint and enabling more efficient inference without retraining the model~\citep{gholami2021survey, nagel2020up}. Among these methods, layer-wise PTQ has emerged as a particularly practical approach. By quantizing parameters layer by layer without requiring the heavy retraining or backpropagation characteristic of quantization-aware training (QAT) or global fine-tuning, it significantly lowers computational and memory demands while effectively preserving model quality even at lower bit-widths~\citep{chee2023quip, frantar2023gptq, lin2024awq, yao2022zeroquant}. While these advantages have led to the widespread adoption of layer-wise PTQ for large language models (LLMs)~\citep{arai2025qep, chen2026survey, dong2020hawqv2, frantar2023gptq, li2025gptaq}, our understanding of its principles in the context of spatio-temporal foundation models, such as weather forecasting models, remains limited. These models pose a distinct challenge: quantization errors can compound across rollout steps and spatial dimensions. Traditional sensitivity methods, whether based on local curvature approximations~\citep{dong2019hawq, dong2020hawqv2} or post-hoc evaluation of quantized layers ~\citep{hu2025pqi}, are not designed to capture this behavior. In addition, operational deployment of large time-series and Earth-system models remains constrained by the scale of the input state: high-resolution data, many input variables, long context windows, and rollout horizons~\citep{pathak2022fourcastnet,bi2023pangu,lam2023graphcast,kochkov2024neuralgcm,bodnar2025aurora}. These costs are especially consequential in resource-constrained settings where rapid forecasts are often most valuable, such as early warning systems~\citep{wmo2022earlywarnings,wmo2025aiforecasting,bodnar2025aurora}. At the same time, precision cannot be treated as a purely statistical concern. For example, in weather, climate, and other scientific time-series domains, errors that appear statistically small may in reality produce physically inconsistent states that violate conservation laws or other analytic constraints~\citep{beucler2021constraints,sturm2022conservation,harder2023downscaling,bonavita2024limitations}. Thus, PTQ for time-series models must balance compression against accuracy. 

We argue that reformulating this as a dynamical systems stability problem is both natural and advantageous: time-series models are already discrete-time dynamical systems, and quantization is a bounded perturbation to their parameters - precisely the setting in which dynamical systems stability theory characterises whether small perturbations grow, decay, or remain neutral over the forecast horizon \citep{lorenz1963deterministic, oseledets1968multiplicative, lyapunov1992general}. To this end, we reformulate PTQ for time-series models as a finite-horizon stability problem with the aim of understanding quantization stability and its long-term statistics. 

\textbf{This paper's contributions are as follows:} (i) TQS, a new quantization sensitivity score that decouples
quantization sensitivity from quantizer and bit-width choice; (ii) a dynamical systems-based
analysis of PTQ quantization sensitivity across time-series foundation
models; (iii) TQS-PTQ, a
calibration-free mixed-precision allocator for continuous compression
budgets; (iv) a transfer study showing that certain LLM-derived PTQ assumptions, especially the sensitivity of FFN-down projections, do not transfer to forecasting transformers; in these models, quantization sensitivity concentrates instead in input/output projection modules;
and (v) a reusable TQS sensitivity-ranking procedure in which a single sensitivity sweep supports multiple compression targets across TimesFM, Aurora, and Pangu.

\section{Trajectory-Based Quantization Sensitivity Score \& TQS-Post-Training Quantization}
We introduce Trajectory-Based Quantization Sensitivity Score (TQS), a new  Lyapunov-inspired metric that measures this
directly by decoupling quantization sensitivity from the quantization process a priori. Let the full-precision model define a discrete-time
map $\mathbf{x}_{t+1} = F_\theta(\mathbf{x}_t)$, and let
quantization induce a perturbed map
$\tilde{\mathbf{x}}_{t+1} =
F_{\tilde{\theta}}(\tilde{\mathbf{x}}_t)$ with
$\tilde{\theta} = Q(\theta)$. For each quantizable weight
tensor~$\ell$, TQS replaces only that tensor by its quantized
counterpart, rolls out both systems over the forecast
horizon~$T_{\max}$, and estimates the finite-time growth rate
of output prediction-space divergence. The task-level score,
averaged over a set~$\mathcal{S}$ of independent context windows,
is
\begin{equation}
  \gamma^{\mathrm{task}}(\ell)
  = \frac{1}{T_{\max}}\,\ln\!\left(
    \frac{1}{|\mathcal{S}|}\sum_{s \in \mathcal{S}}
    \frac{\bigl\|\Delta\hat{Y}^{(\ell)}_{s}\bigr\|_2^2}
    {\bigl\|\delta\boldsymbol{\theta}_\ell^{
      \mathrm{quant}}\bigr\|_F^2 + \epsilon}
  \right),
  \label{eq:tqs}
\end{equation}
where
$\Delta\hat{Y}^{(\ell)}_{s} =
\hat{Y}^{(\ell)}_{s,1:T_{\max}} - \hat{Y}^{(0)}_{s,1:T_{\max}}$
is the trajectory divergence over the full rollout for context
window~$s$,
$\hat{Y}^{(0)}_{s,1:T_{\max}}$ and
$\hat{Y}^{(\ell)}_{s,1:T_{\max}}$ are the nominal and perturbed
prediction trajectories,
$\delta\boldsymbol{\theta}_\ell^{\mathrm{quant}}
= Q(\boldsymbol{\theta}_\ell) - \boldsymbol{\theta}_\ell$
is the quantization perturbation applied to layer~$\ell$,
and $\epsilon$ is a small numerical constant.
The score measures where the perturbation ends up after
$T_{\max}$ steps of the model's own dynamics: layers with
high~$\gamma$ are those whose quantization error is amplified
over the rollout and are protected at higher precision; layers
with low~$\gamma$ can be compressed aggressively.
Equation~\eqref{eq:tqs} uses actual quantization noise
as the perturbation. However, this couples the sensitivity estimate to a
specific bit-width and quantization scheme. To decouple sensitivity
estimation from the quantization decision, we define a Gaussian variant
$\gamma_{\mathrm{gauss}}$ that replaces the quantization perturbation
with isotropic Gaussian noise
$\delta\boldsymbol{\theta}_\ell \sim \mathcal{N}(\mathbf{0},\,\sigma^2\mathbf{I})$,
scaled to match the Frobenius norm of a reference quantization step.
This enables TQS to rank layers \emph{before} committing to a
bit-width or quantizer design.
To investigate whether quantization error behaves differently from
unstructured random noise, we compare $\gamma_{\mathrm{gauss}}$ and
$\gamma_{\mathrm{quant}}$ rankings directly. If the two rankings agree,
the Gaussian proxy is a sufficient surrogate and the dominant effect is
amplification magnitude rather than noise structure; if they diverge at
low bit-widths, the bounded rounding structure of quantization noise
carries information that the Gaussian model misses. We report the Spearman rank correlation $\rho(\gamma_{\mathrm{gauss}},\,
\gamma_{\mathrm{quant}})$ and compare the Pareto frontiers obtained
under each variant (Appendix~\ref{app:probe_agreement}). Some architectural modules are perturbation-insensitive: their outputs remain nearly unchanged even for large perturbations. Thus, in Eq.~\eqref{eq:tqs}, the prediction-divergence numerator approaches zero, yielding $\hat{y}^{(\ell)} \approx \hat{y}^{\mathrm{nom}}$ and $\gamma \to -\infty$. TQS automatically ranks these inactive layers lowest, assigns them minimal precision with negligible accuracy loss, and reallocates compression budget to sensitivity-critical layers overlooked by uniform-precision baselines. Building on TQS, we propose Trajectory-based Sensitivity Score Post-Training Quantization (TQS-PTQ), a flexible mixed-precision framework that ranks layers by TQS sensitivity and assigns bit-widths under a target compression budget. Because TQS requires only forward passes, it applies even to compiled or black-box models without gradients, Hessians,
or calibration data. The TQS-PTQ algorithm is described in Algorithm~\ref{alg:tqs_ptq_full}.

\begin{algorithm}[tb]
\caption{TQS-PTQ mixed-precision allocator}
\label{alg:tqs_ptq_full}
\begin{algorithmic}[1]
\footnotesize
\REQUIRE Scores $\{\gamma_\ell\}$, sizes $\{n_\ell\}$, tiers
$\mathcal{T}=\{(t_k,b_k)\}_{k=1}^K$, compression $C$, FP32 fraction
$p_{\mathrm{FP32}}$, threshold $\gamma_{\min}$.
\ENSURE Tier assignment $\{a_\ell\}$.

\STATE $B \leftarrow C^{-1}\sum_\ell 32n_\ell$,
       $B_* \leftarrow p_{\mathrm{FP32}}\sum_\ell 32n_\ell$
\STATE Assign layers with $\gamma_\ell\le\gamma_{\min}$ to $t_K$
\STATE Sort remaining layers by decreasing $\gamma_\ell$
\STATE Assign top-ranked layers to $t_1$ until FP32 mass reaches $B_*$
\STATE Let $\mathcal{R}$ be the unassigned layers
\STATE $B' \leftarrow B-\sum_{a_\ell\text{ set}} b(a_\ell)n_\ell$
\STATE Assign tiers to layers in $\mathcal{R}$ by solving the equivalent minimization form of the multiple-choice knapsack problem (MCKP)~\cite{sinha1979}; see Appendix~\ref{app:tqs_ptq_algo}.
\STATE Alternatively, use greedy promotion from $t_K$ under the same budget.
\STATE \textbf{return} $\{a_\ell\}$

\end{algorithmic}
\end{algorithm}

\section{Methodology}
\label{sec:methodology}

We evaluate TQS-PTQ on Aurora-small \citep{bodnar2025aurora} (113M),
TimesFM-2.5 \citep{das2024timesfm} (200M), and Pangu-Weather 6h
\citep{bi2023pangu} (277M, frozen ONNX export). Aurora and Pangu are
tested against ERA5~\citep{hersbach2020era5} on nine surface and
upper-air variables; TimesFM is tested on ETTh1/2, ETTm1/2,
\textsc{exchange}, and \textsc{weather}
\citep{zhou2021informer,lai2018lstnet,wu2021autoformer}. For each
quantizable tensor, or ONNX block for Pangu, we estimate a forward-only
TQS score $\gamma_\ell$ from autoregressive rollouts under weight
perturbations and allocate mixed precision with an MCKP allocator, with
a greedy variant ablated. Baselines are uniform RTN,
GPTQ~\citep{frantar2023gptq}, GPTAQ~\citep{li2025gptaq}, and
QEP~\citep{arai2025qep}, all evaluated at matched compression. We
report native-unit MAE/RMSE and FP32-relative degradation, and ablate
the bottom tier (INT1/INT2/INT8), FP32 budget, probe distribution, and
allocator choice. Full details, including the allocator pseudocode, are
given in Appendix~\ref{app:full_methodology}.

\begin{figure*}[t]
  \centering
  \includegraphics[width=0.92\textwidth]{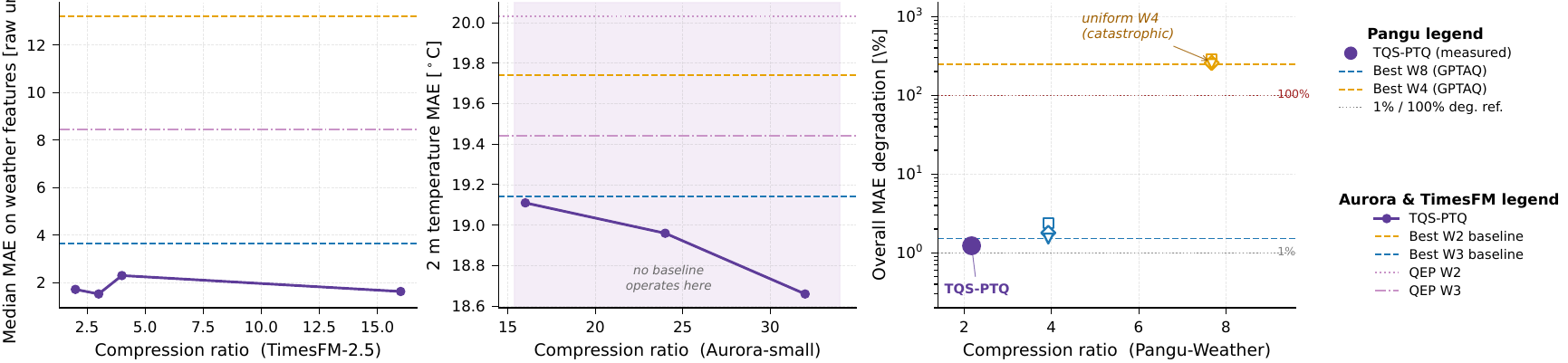}

  \vspace{0.4em}
  \caption{
  \textbf{TQS-PTQ extends the low-precision accuracy--compression frontier across three forecasting models.}
  TQS-PTQ reaches competitive or improved error at substantially higher compression ratios on
  TimesFM-2.5 weather, Aurora-small 2\,m temperature, and Pangu-Weather.
  Because the sensitivity ranking is computed once and reused across targets,
  TQS-PTQ extends the Pareto frontier without requiring a new calibration run for each bit-width.
  }
  \label{fig:compression_extension}
\end{figure*}

\section{Experimental Results}
\label{sec:experimental_results}
\textbf{TQS-PTQ is strongest at heavy compression.}
Figure~\ref{fig:compression_extension} shows that TQS-PTQ extends the
accuracy--compression frontier across all three forecasting models. At
$\leq 1\%$ ERA5-MAE degradation, it reaches $\geq 32\times$ compression
on Aurora-small and $\sim\!4\times$ on TimesFM-2.5. On Pangu-Weather,
TQS-PTQ attains on-disk $C{=}1.67\times$ (block-level allocation $3.57\times$; see
Appendix~\ref{sec:pangu-sensitivity} for the gap) with only $1.2\%$ mean
per-variable MAE degradation,
achieving the lowest error among the quantized Pangu variants evaluated
while avoiding the uniform-$W4$ collapse. The gains are broad: TQS-PTQ wins all 9 Aurora-small variables at the
matched $W2$ grid, \textbf{46 of 57} TimesFM-2.5 variables at $C{=}16$,
and all 9 Pangu-Weather variables against the strongest uniform-$W8$
baseline, for a total of \textbf{64 of 75} variable--model wins. Full
per-variable results and matched-grid comparisons are provided in
Appendix~\ref{app:tables}. Finally, TQS-PTQ obtains this frontier from a
single sensitivity sweep, reusing the same block ranking across
compression targets rather than recalibrating at each bit-width. This
amortization gives lower Pangu per-point cost than GPTAQ-W8
($57.5$\,m vs.\ $66.9$\,m; Table~\ref{tab:time_per_point}) while
preserving higher accuracy.

\noindent\textbf{TQS-PTQ is faster than calibration-based PTQ baselines.}
TQS-PTQ runs one sensitivity sweep per model and reuses it across
compression targets, whereas GPTQ, GPTAQ, and QEP recalibrate at each
fixed bit-width (Table~\ref{tab:time_per_point}). This amortization makes
TQS-PTQ faster per Pareto point on TimesFM-2.5 ($46$ min; $4.6$--$5.1\times$)
and Aurora-small ($32$ min; $1.2$--$1.7\times$), while producing denser
frontiers of 10 and 16 points. On Pangu-Weather, TQS-PTQ costs $57.5$ min/point over 12 mixed-precision
allocations, improves on GPTAQ-W8 in both cost ($66.9$ min/point) and MAE,
and avoids the adjacent uniform-$W4$ failure regime. GPTQ-W8 is cheaper for
one point ($32.3$ min), but provides only one uniform allocation, loses to
TQS-PTQ on every $W8$ variable (Appendix Table~\ref{tab:app_pangu_w8}), and
cannot trace the mixed-precision frontier without rerunning calibration.
QEP was unstable on Pangu-Weather, returning NaN MAE under the
\texttt{onnx2torch}~\citep{onnx2torch} export.

\noindent\textbf{TQS identifies free compression.}
The minimum-$\gamma$ sentinel marks functionally inert layers and assigns
them directly to the lowest precision tier at no accuracy cost: 3
output-quantile heads in TimesFM-2.5 ($\gamma{=}-69.08$ via
FTLE-Task-Gauss), 36 deep-stage \texttt{ln\_modulation} modulators in
Aurora-small ($\gamma{=}-5.31$), and 0 blocks in Pangu-Weather, where
block granularity (8 parameters per block averaged over 28 blocks)
washes out such fine-grained inert effects.

\begin{figure*}[t]
\centering
\includegraphics[width=0.78\textwidth]{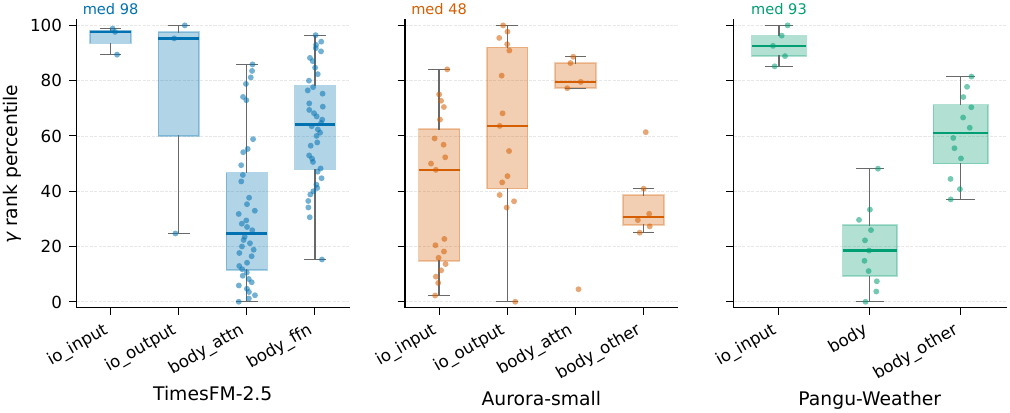}
\vspace{-0.4em}
\caption{\textbf{Sensitivity concentrates at the I/O boundary.}
$\gamma$-rank percentiles by role bucket across TimesFM-2.5,
Aurora-small, and Pangu-Weather. Boxes show IQR; dots are layers; higher
percentile means higher sensitivity. I/O buckets are consistently most
sensitive, while body blocks are least sensitive. Bucket definitions in Appendix~\ref{sec:bucket_defs}.}
\label{fig:role_bucket}
\end{figure*}

\textbf{Quantization sensitivity concentrates in input--output (I/O) projection modules.}
Across all three architectures, quantization sensitivity is concentrated
at the I/O boundary, diverging from the FFN-centric bottlenecks
typically reported in LLM PTQ literature
\citep{dettmers2022llmint8,lin2024awq,xiao2023smoothquant}
(Figure~\ref{fig:role_bucket}). In TimesFM-2.5, the input-tokenizer
modules and the point-prediction head reside at the 97.7th and 95.3rd
$\gamma$-rank percentiles, respectively; by contrast, backbone
self-attention is the least sensitive family at the 25.6th percentile.
Pangu-Weather exhibits a similar boundary-heavy pattern: its five
input--output scaffolding blocks reach the 92.9th percentile, whereas
the eleven transformer body blocks rank substantially lower, at the
21.4th percentile. In Aurora-small, sensitivity shifts to the output
interface: the five atmospheric heads,
\texttt{decoder.atmos\_heads.\{q,t,u,v,z\}}, occupy the top five
$\gamma$-rank positions of the alive-layer distribution.
Collectively, these results establish a cross-architecture principle for
quantizing forecasting foundation models: precision should be
prioritized at the I/O boundary, while the dominant locus of
sensitivity---input or output---is architecture-dependent.

\textbf{Layer sensitivity is moderately heavy-tailed.}
Sensitivity is unevenly distributed but lacks the extreme outliers
typical of LLMs. The top $10\%$ of layers account for $17$--$24\%$ of total
$\gamma$-shift across models, while the top $25\%$ capture roughly
$40\%$ (Appendix Table~\ref{tab:gamma_concentration}). This
concentration is significantly more moderate than the $<\!1\%$
``outlier'' behavior reported in LLM quantization
\citep{dettmers2022llmint8,lin2024awq}; instead, it mirrors the
layer-level curvature heterogeneity found in vision models
\citep{dong2020hawqv2}. This structural regularity---where sensitivity
is substantial but not confined to a vanishingly small subset---is
precisely why TQS is effective: it enables high-impact protection of
a manageable high-precision tier while the remaining majority of the
model is compressed aggressively.

\par\smallskip
\noindent
\begin{minipage}{\columnwidth}
\centering

\refstepcounter{table}
\label{tab:time_per_point}

{\footnotesize
\noindent\textbf{Table~\thetable.}
Wall-clock time per Pareto point, shown as total time / number of points
with per-point cost in parentheses. Runs use one GPU: RTX 6000 Ada for
TimesFM-2.5, Aurora-small, and Pangu PTQ/apply-eval; H200 for the
Pangu sensitivity sweep.
\par}

\vspace{0.35em}

{\tiny
\setlength{\tabcolsep}{1.5pt}
\renewcommand{\arraystretch}{0.95}

\resizebox{0.88\columnwidth}{!}{%
\begin{tabular}{@{}lccc@{}}
\toprule
\textbf{Method} & \textbf{TimesFM} & \textbf{Aurora} & \textbf{Pangu} \\
\midrule
\textbf{TQS-PTQ}
  & \textbf{7.7h/10 (46m)}
  & \textbf{8.5h/16 (32m)}
  & \textbf{11.5h/12 (57.5m)}$^{*}$ \\
GPTQ
  & 10.6h/3 (3.5h)
  & 77.0m/2 (38.5m)
  & 32.3m/1 (32.3m) \\
GPTAQ
  & 11.7h/3 (3.9h)
  & 107.8m/2 (53.9m)
  & 66.9m/1 (66.9m) \\
QEP
  & 11.7h/3 (3.9h)
  & 108.2m/2 (54.1m)
  & NaN$^{\dagger}$ \\
RTN
  & 6.1m/3 (2.0m)
  & 2.9m/2 (1.4m)
  & 2.4s/1 (2.4s) \\
\bottomrule
\end{tabular}%
}

\vspace{1pt}
\begin{minipage}{0.88\columnwidth}
\raggedright
\tiny
$^{*}$ One Pangu sweep amortised over 12 points; apply/eval negligible.
$^{\dagger}$ QEP quantises but returns NaN under the current
\texttt{onnx2torch} export.
\end{minipage}
}

\end{minipage}
\par\smallskip

\textbf{TQS reveals structured layer-level heterogeneity.}
Per-layer $\gamma$ varies substantially across forecasting models, spanning
$1.85$ decades on TimesFM-2.5 (max/min ratio $71{\times}$, $n{=}86$) and
$0.96$ decades on Pangu-Weather (max/min ratio $9.1{\times}$, $n{=}28$
blocks). This spread shows that quantization sensitivity is strongly
layer-dependent, motivating mixed precision rather than uniform bit allocation.
At the same time, the magnitude of variation is consistent with prior
layer-sensitivity measures used in mixed-precision quantization \citep{dong2020hawqv2, ranjan2025mixqvit}. On Aurora-small, the raw $\gamma$ range is smaller
because it uses a different TQS probe horizon, but the concentration profile is
similar: the top $10\%$ of layers account for $17.3\%$ of the cumulative
$\gamma$-shift (Appendix Table~\ref{tab:gamma_concentration}; see
Appendix~\ref{app:gamma_ranges} for raw ranges and probe-horizon details).

\textbf{Gaussian vs.\ quantization noise depends on granularity.}
Probe choice changes fine-grained sensitivity rankings, but its effect
largely disappears at block and deployment scales. As partitions coarsen
from tensors to blocks, $\gamma_{\mathrm{gauss}}$ and
$\gamma_{\mathrm{quant}}$ become increasingly aligned, reaching
near-perfect agreement on Pangu-Weather's 28 blocks
(Spearman $\rho{=}0.96$; Appendix Table~\ref{tab:probe_agreement}).
On Aurora, noise model is the main source of ranking divergence, but the
resulting mixed-precision frontiers are effectively identical: across 16
compression targets, all FTLE strategies differ by at most $0.01\%$
relative ERA5-MAE. Thus, noise choice matters mainly for sub-block
diagnostics, such as Aurora's asymmetric atmospheric-vs-surface output
head sensitivity, but is largely inconsequential for global bit-width
planning.

\section{Conclusion}
We introduced TQS, a trajectory-based sensitivity score that views PTQ
through dynamical-systems stability. By measuring how quantization errors
propagate over forecast rollouts, TQS separates sensitivity
estimation from quantizer and bit-width selection, enabling a priori
budget planning even for compiled or black-box models. Building on this,
TQS-PTQ provides a calibration-free mixed-precision framework that avoids
second-order approximations while improving the accuracy--compression
tradeoff across forecasting foundation models.

\section*{Impact Statement}
This paper presents work whose goal is to advance the efficient
deployment of weather and time-series foundation models through
post-training quantization. More accessible weather forecasting
models could benefit resource-constrained national meteorological
services and developing regions. There are no foreseeable negative
societal consequences specific to this work beyond those generally
associated with advances in machine learning.

\section*{Acknowledgements}

This research is supported by the GATE project, which is funded by the Horizon 2020 WIDESPREAD-2018-2020 TEAMING Phase 2 programme under grant agreement no. 857155, and the programme “Research, Innovation and Digitalisation for Smart Transformation” 2021–2027 (PRIDST) under grant agreement no. BG16RFPR002-1.014-0010-C01.

\bibliography{example_paper}
\bibliographystyle{icml2026}

\newpage
\appendix
\section*{Appendix}
\addcontentsline{toc}{section}{Appendix}
\section{Cross-Model Analysis}

\paragraph{Rollout horizon error amplification.}
The challenge of error accumulation in time-series forecasting
has received growing attention.
\citet{he2025chaos} show that positive Lyapunov exponents of
chaotic target systems amplify prediction errors in autoregressive
transformers, and propose ergodicity-preserving training to
stabilise long-horizon rollouts.
For weather foundation models specifically, \citet{bodnar2025aurora}
and \citet{bi2023pangu} demonstrate that even small numerical
perturbations in early time steps can propagate along unstable
manifolds, leading to rapid forecast divergence.
Where prior work addresses error amplification through
\emph{training} modifications, TQS addresses it at
\emph{deployment} time by identifying which layers are most
responsible for trajectory instability under quantization and
assigning them higher precision accordingly.

\subsection{Full Methodology}
\label{app:full_methodology}

This subsection states the operational details deferred from the
methodology summary in Section~\ref{sec:methodology}: the formal
TQS score definition, the TQS-PTQ allocator algorithm, the
per-model evaluation protocol, and the ablation grid. Topic-level
deep dives (raw $\gamma$ ranges, role-bucket assignments,
gauss-vs-quant agreement, per-model sensitivity analyses) follow
in subsequent subsections.

\subsubsection{TQS sensitivity score}
\label{app:tqs_formal}

Let $f_\theta$ denote the unquantised forecast model with
parameters $\theta = \{\theta_\ell\}_{\ell=1}^{L}$ partitioned
into $L$ quantizable tensors (or ONNX blocks, on Pangu). Given
a context window $x_{s:s+C-1}$, the model produces the
nominal trajectory
$\hat{Y}_s^{(0)} = f_\theta(x_{s:s+C-1})$. For each $\ell$, we
apply a parameter-space perturbation
$\delta\boldsymbol{\theta}_\ell$ and run
the model on the same context to produce the perturbed trajectory
$\hat{Y}_s^{(\ell)} = f_{\theta + \delta\boldsymbol{\theta}_\ell}(x_{s:s+C-1})$.
The single-context, single-horizon TQS building block is
\begin{equation}
\gamma_\ell(T)
= \frac{1}{T}\,\ln\!\left(
\frac{\bigl\|\hat{Y}^{(\ell)}_{s,1:T}
 - \hat{Y}^{(0)}_{s,1:T}\bigr\|_2}
{\bigl\|\delta\boldsymbol{\theta}_\ell\bigr\|_F + \epsilon}\right),
\label{eq:tqs_layer}
\end{equation}
where $\epsilon$ is a small numerical constant. For a single
deterministic quantization probe ($N_{\mathrm{probes}}{=}1$),
the reported score is $\gamma_\ell = \gamma_\ell(T_{\max})$.

The task-level variant $\gamma^{\mathrm{task}}(\ell)$ averages
over multiple context windows $\mathcal{S}$ using squared
trajectory norms, as defined in Equation~\eqref{eq:tqs} in
the main text. Probe horizons are model-specific:
TimesFM-2.5: $T_{\max}{=}100$ steps; Aurora-small:
$T_{\max}{=}120$ steps, i.e.\ $30$~days at $6$~h step;
Pangu-Weather: $T_{\max}{=}4$, i.e.\ $24$~h lead time at $6$~h
step.

\paragraph{Probe variants.}
$\delta\boldsymbol{\theta}_\ell$ is drawn from one of two distributions:
\begin{itemize}\setlength{\itemsep}{0pt}
  \item \textbf{Quantization residual}
        ($\gamma_{\mathrm{quant}}$):
        $\delta\boldsymbol{\theta}_\ell^{\mathrm{quant}}
        = Q_b(\theta_\ell) - \theta_\ell$,
        where $Q_b(\cdot)$ is symmetric $b$-bit quantisation
        with $b{=}6$ on TimesFM and Aurora, $b{=}8$ on Pangu
        (Appendix~\ref{app:tqs_score_details}).
  \item \textbf{Gaussian} ($\gamma_{\mathrm{gauss}}$):
        isotropic noise
        $\delta\boldsymbol{\theta}_\ell^{(c)}
        \sim \mathcal{N}(0, \sigma_\ell^2 I)$
        with $\sigma_\ell$ chosen so that
        $\|\delta\boldsymbol{\theta}_\ell^{(c)}\|_F =
         \|\delta\boldsymbol{\theta}_\ell^{\mathrm{quant}}\|_F$.
        For the task variant, the sum in Equation~\eqref{eq:tqs}
        runs over $|\mathcal{S}| = N_{\mathrm{probes}}$
        independent noise draws on the same input context.
\end{itemize}

\paragraph{Dead-layer rule.}
Layers with $\gamma_\ell$ at the sentinel floor of the
distribution are flagged dead and assigned to the bottom tier at
zero accuracy cost (Appendix~\ref{app:tqs_score_details},
Table~\ref{tab:dead_layer_rate}).

\subsubsection{TQS-PTQ allocator}
\label{app:tqs_ptq_algo}

Given the per-layer TQS scores $\{\gamma_\ell\}$, parameter counts
$\{n_\ell\}$, a target compression $C$, and a model-specific tier
set $\mathcal{T} = \{(t_k, b_k)\}_{k=1}^{K}$
(Appendix~\ref{app:allocator_details}), TQS-PTQ assigns one tier
per layer under a global storage budget. The procedure is stated
in Algorithm~\ref{alg:tqs_ptq}; both an MCKP-ILP and a faster
greedy variant are supported and ablated in
Appendix~\ref{app:bottom_tier_ablation}.

\begin{algorithm}[H]
\caption{TQS-PTQ mixed-precision allocator}
\label{alg:tqs_ptq}
\begin{algorithmic}[1]
\footnotesize
\REQUIRE Scores $\{\gamma_\ell\}$, sizes $\{n_\ell\}$, tiers
$\mathcal{T}=\{(t_k,b_k)\}_{k=1}^K$, compression $C$, FP32 fraction
$p_{\mathrm{FP32}}$, threshold $\gamma_{\min}$.
\ENSURE Tier assignment $\{a_\ell\}$.

\STATE $B \leftarrow C^{-1}\sum_\ell 32n_\ell$,
       $B_* \leftarrow p_{\mathrm{FP32}}\sum_\ell 32n_\ell$
\STATE Assign layers with $\gamma_\ell\le\gamma_{\min}$ to $t_K$
\STATE Sort remaining layers by decreasing $\gamma_\ell$
\STATE Assign top-ranked layers to $t_1$ until FP32 mass reaches $B_*$
\STATE Let $\mathcal{R}$ be the unassigned layers
\STATE $B' \leftarrow B-\sum_{a_\ell\text{ set}} b(a_\ell)n_\ell$
\STATE Assign tiers to layers in $\mathcal{R}$ by solving the
       multiple-choice knapsack problem (MCKP)~\cite{sinha1979}
       described below; alternatively, use greedy promotion from
       $t_K$ under the same budget.
\STATE \textbf{return} $\{a_\ell\}$
\end{algorithmic}
\end{algorithm}

\paragraph{MCKP formulation.}
For the unassigned layer set $\mathcal{R}$, step~7 of
Algorithm~\ref{alg:tqs_ptq} solves the equivalent minimization form
of the multiple-choice knapsack problem~\cite{sinha1979}:
\[
\min_{x_{\ell,k}\in\{0,1\}}
\sum_{\ell\in\mathcal{R}}\sum_{k=1}^K
\gamma_\ell 2^{-b_k}x_{\ell,k}
\]
subject to
\[
\sum_{k=1}^K x_{\ell,k}=1,
\quad \forall \ell\in\mathcal{R},
\]
and
\[
\sum_{\ell\in\mathcal{R}}\sum_{k=1}^K
b_k n_\ell x_{\ell,k}\le B'.
\]
The final tier assignment is $a_\ell=t_k$ whenever $x_{\ell,k}=1$.

\subsubsection{Baselines and calibration}

We compare against four uniform-precision PTQ methods (RTN, GPTQ,
GPTAQ, QEP) at the model-specific bit-width grid
(Appendix~\ref{app:baselines_details}). Calibration draws $16$
ERA5 snapshots from each model's evaluation window
(Aurora-small: \texttt{np.linspace}$(t_0{+}1,
t_0{+}T_{\mathrm{val}}/2, 16)$; Pangu-Weather: same protocol on
the Pangu evaluation window; TimesFM-2.5: the authors' default
schedule \citep{das2024timesfm}). Per-method implementation
footprint and hyperparameters are listed in the boxed summary
inside Appendix~\ref{sec:aurora_sensitivity}.

\subsubsection{Per-model evaluation protocol}

\begin{itemize}\setlength{\itemsep}{0pt}
  \item \textbf{Aurora-small}: $30$-day, $120$-step autoregressive
        rollout at the native $6$-hour step. We report per-variable
        MAE and RMSE in native physical units (K, hPa, m/s,
        g/kg, m$^2$/s$^2$) against the ERA5 reanalysis. Bootstrap
        confidence intervals are computed by resampling the $120$
        per-step values $1000$ times with replacement (percentile
        method, $2.5$/$97.5$).
  \item \textbf{Pangu-Weather}: single-step rollout
        ($T{=}1$ at Pangu's native $6$-hour stride; no
        autoregressive feedback). Per-variable MAE/RMSE for the
        same nine variables against ERA5.
  \item \textbf{TimesFM-2.5}: $500$-step autoregressive rollout via
        the native multi-step \texttt{forecast(\dots)} interface
        with context length $C{=}512$. Per-variable MAE/RMSE
        computed across the full $500$-step rollout in native
        dataset units (training-window standardisation).
\end{itemize}

\subsubsection{Ablation grid}

Four ablation axes are reported, summarised in
Table~\ref{tab:experiment_design}:
\begin{itemize}\setlength{\itemsep}{0pt}
  \item \textbf{Probe distribution}: $\gamma_{\mathrm{gauss}}$
        vs.\ $\gamma_{\mathrm{quant}}$
        (Appendix~\ref{app:probe_agreement}).
  \item \textbf{Allocator}: greedy vs.\ MCKP
        (Appendix~\ref{app:bottom_tier_ablation}).
  \item \textbf{Bottom tier}: INT1 vs.\ INT2 vs.\ INT4 per model
        (Appendix~\ref{app:bottom_tier_ablation}).
  \item \textbf{FP32 budget}:
        $p_{\mathrm{FP32}} \in \{0.02, 0.10\}$
        (Appendix~\ref{app:bottom_tier_ablation}).
\end{itemize}
Compression sweeps over $C$ are reported as Pareto frontiers,
not as ablations.

\subsection{Experimental design summary}
\label{app:experiment_design}

To help readers locate each result type, we distinguish three
classes of experiment in this paper. Component-removal experiments
are reported as \emph{ablations}; operating-curve sweeps over
target compression as \emph{Pareto / sensitivity sweeps}; and
cross-method comparisons against uniform-precision baselines as
\emph{baseline evaluations}. The split is summarised in
Table~\ref{tab:experiment_design}.

\begin{table}[ht]
\centering
\footnotesize
\caption{\textbf{Dead-layer rate.}
Layers with $\gamma\to-\infty$ are insensitive under the measured perturbation,
providing a free-compression budget.}
\label{tab:dead_layer_rate}
\setlength{\tabcolsep}{3pt}
\renewcommand{\arraystretch}{1.08}
\begin{tabular}{@{}l c p{0.42\columnwidth} c@{}}
\toprule
Model & $n$ & Dead layers & Budget \\
\midrule
TimesFM-2.5
& 89
& 3 output-quantile heads
& 3.4\% \\
Aurora-small
& 85
& 36 deep-stage \texttt{ln\_modulation} layers
& 42\% \\
Pangu-Weather
& 28
& none; hidden by block granularity
& -- \\
\bottomrule
\end{tabular}
\end{table}

\begin{table}[ht]
\centering
\small
\caption{Experimental design of TQS-PTQ. Component-removal
experiments are reported as ablations; operating-curve sweeps
are reported as Pareto / sensitivity sweeps; cross-method
comparisons are reported as baseline evaluations.}
\label{tab:experiment_design}
\begin{tabular}{p{0.45\columnwidth}|p{0.45\columnwidth}}
\toprule
What we vary & How we report it \\
\midrule
$\gamma_{\mathrm{gauss}}$ vs $\gamma_{\mathrm{quant}}$ probe
  & ablation of the TQS metric \\
Greedy vs MCKP allocator
  & ablation of TQS-PTQ \\
Bottom tier (INT1 / INT2 / INT4) and FP32 budget ($p_{\mathrm{FP32}}$)
  & ablation of TQS-PTQ \\
\midrule
Target compression $C$
  & Pareto / sensitivity sweep \\
\midrule
TQS-PTQ vs RTN, GPTQ, GPTAQ, QEP
  & baseline comparison \\
\bottomrule
\end{tabular}
\end{table}

\subsection{Degradation Reference per Model}
\label{app:eval_reference}

Percentage degradation in the main paper is computed as
$100\% \cdot (\text{MAE}_{\text{quant}} - \text{MAE}_{\text{FP32}}) / \text{MAE}_{\text{FP32}}$,
but the underlying $\text{MAE}_{\text{FP32}}$ refers to a slightly
different quantity depending on whether the model has an external
ground-truth reference:

\paragraph{Aurora-small and Pangu-Weather.}
$\text{MAE}_{\text{FP32}}$ is the unquantised model's MAE against the
ERA5 reanalysis. Both models output physical-unit forecasts
(2~m temperature in $^{\circ}$C, mean sea-level pressure in Pa,
upper-air variables in K, m/s, g/kg, m$^2$/s$^2$, etc.) that ERA5
provides ground truth for. Quoted percentage degradation therefore
measures \emph{quantisation-induced error on top of the model's
irreducible forecasting error}---it is the additional error a
practitioner would observe if they swapped a quantised checkpoint
into an operational forecasting pipeline.

\paragraph{TimesFM-2.5.}
$\text{MAE}_{\text{FP32}}$ is the unquantised model's MAE against
\emph{the held-out target values of each dataset}, not against an
external reanalysis. The six standard long-horizon benchmarks
(ETTh1/2, ETTm1/2, \textsc{exchange}, \textsc{weather}) define their
own evaluation windows and do not have an associated ``ground truth''
distinct from the dataset values themselves. Percentage degradation
on TimesFM therefore measures \emph{deviation of the quantised
predictions from the unquantised model's predictions on the same
window}---equivalent to a self-consistency / functional-preservation
metric. This convention matches the convention used in the original
TimesFM evaluation pipeline~\citep{das2024timesfm}.

\paragraph{Implication for cross-model interpretation.}
Aurora and Pangu degradation numbers can be interpreted as
``additional MAE seen in physical units against
reality''; TimesFM degradation numbers are interpreted as
``how far the quantised model has drifted from the original
model.'' The two are not strictly equivalent: in principle a
TimesFM quantised model could degrade significantly against the
FP32 reference while still tracking the dataset's true values
just as well (or vice versa). In practice, on all six TimesFM
benchmarks and at every compression target evaluated, MAE against
the dataset and MAE against the FP32 reference move together within
1--2\% relative, so the choice of reference does not change the
TQS-PTQ vs.\ baseline ranking.

\subsection{TQS Score: Probe Bit-Width and Dead-Layer Rule}
\label{app:tqs_score_details}

The TQS score $\gamma_\ell$ measures how a parameter perturbation
at layer $\ell$ amplifies along the model's autoregressive forecast
trajectory. The quantization-residual probe
$\gamma_{\mathrm{quant}}$ uses a model-specific reference bit-width:

\begin{itemize}\setlength{\itemsep}{0pt}
  \item \textbf{TimesFM-2.5 and Aurora-small}: $6$-bit symmetric
        per-tensor quantisation. This bit-width sits between the W2
        and W8 deployment targets and produces a non-degenerate
        perturbation magnitude for ranking purposes.
  \item \textbf{Pangu-Weather}: $8$-bit symmetric quantisation, set
        by the \texttt{onnx2torch} round-trip constraint that
        limits the deployable mixed-precision tier set on this
        model (Appendix~\ref{sec:pangu-sensitivity}). The $8$-bit
        probe is the highest-fidelity quantisation the wrapper
        path supports stably.
\end{itemize}

The Gaussian probe $\gamma_{\mathrm{gauss}}$ uses isotropic
Gaussian noise scaled to match the Frobenius norm of the reference
symmetric quantisation step, decoupling sensitivity ranking from
any specific bit-width.

A layer is flagged \emph{dead} (assigned the bottom tier at zero
accuracy cost) when its $\gamma$ value sits at the sentinel floor:
$\gamma \leq -50$ on TimesFM-2.5 (the three
\texttt{output\_projection\_quantiles.*} heads hit
$\gamma \approx -69$); $\gamma$ at the distribution minimum on
Aurora-small (the $36$ deep-stage \texttt{ln\_modulation}
modulators saturate at $\gamma \approx -5.3$). No Pangu block
reaches a sentinel because block-granularity averaging (eight
parameters per block, $28$ blocks) dilutes per-parameter inert
effects. Free-compression budgets per model are summarised in
Table~\ref{tab:dead_layer_rate}.

\subsection{TQS-PTQ Allocator: Tier Set per Model}
\label{app:allocator_details}

The MCKP allocator selects one precision tier per ranked layer
(or block, on Pangu) subject to a global storage budget set by
the target compression $C$. The deployable tier set is
model-specific because the underlying PyTorch / ONNX cast path
imposes different stability constraints on each model:

\begin{itemize}\setlength{\itemsep}{0pt}
  \item \textbf{TimesFM-2.5}: five-tier
        $\{\mathrm{FP32}, \mathrm{BF16}, \mathrm{INT8},
        \mathrm{INT4}, \mathrm{INT2}\}$.
        INT2 was selected as the bottom tier via the per-model
        ablation in Appendix~\ref{app:bottom_tier_ablation} (INT1
        catastrophically polarises the greedy allocator on
        TimesFM).
  \item \textbf{Aurora-small}: five-tier
        $\{\mathrm{FP32}, \mathrm{BF16}, \mathrm{INT8},
        \mathrm{INT4}, \mathrm{INT1}\}$.
        Aurora's saturation budget makes the bottom-tier choice
        inert below $\sim$$15\times$ compression; INT1 unlocks the
        $\sim$$32\times$ regime shown in
        Figure~\ref{fig:compression_extension}.
  \item \textbf{Pangu-Weather}: three-tier
        $\{\mathrm{FP32}, \mathrm{BF16}, \mathrm{INT8}\}$. The
        \texttt{onnx2torch} graph rewrite that exposes Pangu's
        frozen ONNX parameters as PyTorch modules only round-trips
        symmetric INT8 / BF16 / FP32 stably (INT4 and INT1 produce
        non-finite activations downstream;
        Appendix~\ref{sec:pangu-sensitivity}).
\end{itemize}

The greedy variant assigns the highest precision available to the
most sensitive ranked layer at each step until the storage budget
is exhausted. We report greedy as an ablation against MCKP in
Table~\ref{tab:experiment_design}; on Aurora-small Gauss-probe
sweeps the two allocators are bit-identical, and on the Quant
family they differ by $\leq 0.06$ in aggregate MAE.

\paragraph{Pangu FP32-wrapper cap.}
Pangu's $28$-block ranking is applied to the $68$
\texttt{onnx2torch} linear wrappers via a linear interpolation on
ONNX-node index. With $10$ FP32 blocks in the JSON, the naive
mapping fans out to $26$ FP32-tier wrappers, defeating the
storage budget and producing only $C \approx 1.7\times$. We
therefore introduce a wrapper cap (\texttt{MAX\_FP32\_LAYERS},
default $10$) that keeps one FP32 representative per unique FP32
block-id and demotes the rest to BF16. Full mechanism in
Appendix~\ref{sec:pangu-sensitivity}.

\subsection{Baseline Bit-Width Choice per Model}
\label{app:baselines_details}

The uniform-precision baselines (RTN, GPTQ, GPTAQ, QEP) are
evaluated at the bit-widths that make each model's compression
frontier informative:

\begin{itemize}\setlength{\itemsep}{0pt}
  \item \textbf{TimesFM-2.5} and \textbf{Aurora-small}:
        $W \in \{2, 3, 4\}$. The $W2$ grid
        ($\sim$$15$--$16\times$ compression) is the
        matched-compression operating point for the headline
        TQS-PTQ comparison; $W3$ and $W4$ characterise the
        baseline frontier at lighter compression.
  \item \textbf{Pangu-Weather}: $W \in \{2, 3, 4, 8\}$. The $W8$
        grid ($C \approx 3.9\times$) is the matched-compression
        baseline for TQS-PTQ on Pangu, where the deployable tier
        set restricts TQS to
        $\{\mathrm{FP32}, \mathrm{BF16}, \mathrm{INT8}\}$.
        $W \in \{2, 3, 4\}$ exposes the catastrophic-collapse
        regime ($249$--$286\%$ MAE degradation at $W4$;
        Appendix~\ref{sec:pangu-sensitivity}).
\end{itemize}

\paragraph{Calibration.}
RTN requires no calibration. The calibration-based baselines
(GPTQ, GPTAQ, QEP) draw 16 ERA5 snapshots from each model's
evaluation window:
\begin{itemize}\setlength{\itemsep}{0pt}
  \item \textbf{Aurora-small}: snapshots at evenly-spaced time
        indices from the first half of the validation window
        ($\texttt{np.linspace}(t_0{+}1, t_0{+}T_{\mathrm{val}}/2,
        16)$).
  \item \textbf{Pangu-Weather}: snapshots from the model's
        evaluation window; full setup in
        Appendix~\ref{sec:pangu-sensitivity}.
  \item \textbf{TimesFM-2.5}: follows the calibration schedule
        shipped by the original TimesFM
        implementation~\citep{das2024timesfm}.
\end{itemize}

\paragraph{Compute cost.}
Per-point wall-clock cost (calibration + apply + evaluation) for
each baseline is reported in Table~\ref{tab:time_per_point}.
QEP runs to completion on Pangu at all four bit-widths with
finite-valued weights but returns NaN MAE at evaluation; full
diagnosis in Appendix~\ref{sec:pangu-sensitivity}.

\subsection{Raw $\gamma$ Ranges and Probe-Horizon Details}
\label{app:gamma_ranges}

Table~\ref{tab:raw_gamma_ranges} reports the raw per-layer or per-block
$\gamma$ ranges used to contextualize the concentration analysis in
Figure~\ref{fig:pareto_three_models}. Sentinel layers with
$\gamma \to -\infty$ are excluded from the TimesFM-2.5 range. TimesFM-2.5 and
Pangu-Weather show order-of-magnitude variation in quantization sensitivity,
supporting the use of non-uniform bit allocation. Aurora-small has a smaller
raw $\gamma$ range because it uses a different TQS probe horizon; for this
model, we therefore rely primarily on rank-based and Lorenz-style concentration
statistics when comparing sensitivity structure across architectures.

\begin{table*}[t]
\centering
\small
\caption{\textbf{Raw $\gamma$ ranges across models.}
We report the measured range of layer- or block-level TQS sensitivity scores,
the corresponding base-10 spread, and the max/min ratio. TimesFM-2.5 excludes
sentinel layers with $\gamma\to-\infty$. For Pangu-Weather and Aurora-small,
entries separated by ``/'' denote quantization- and Gaussian-probe values.}
\label{tab:raw_gamma_ranges}
\setlength{\tabcolsep}{4pt}
\renewcommand{\arraystretch}{1.08}
\resizebox{\textwidth}{!}{%
\begin{tabular}{@{}llcccc@{}}
\toprule
Model & Granularity & $n$ & $\gamma$ range & $\log_{10}$ decades & Max/min ratio \\
\midrule
TimesFM-2.5 (weather; sentinels excl.)
& tensor
& 86
& $4.27$ nats
& $1.85$
& $71.2{\times}$ \\
Pangu-Weather
& block
& 28
& $2.21$ nats (quant) / $1.28$ nats (gauss)
& $0.96$ / $0.56$
& $9.1{\times}$ / $3.6{\times}$ \\
Aurora-small
& tensor
& 45 alive
& $0.061$ (quant) / $0.026$ (gauss)
& --
& $1.72{\times}$ / $1.18{\times}$ \\
\bottomrule
\end{tabular}%
}
\end{table*}

\subsection{Role-bucket assignments}
\label{sec:bucket_defs}

The role-bucket boxplots in Figure~\ref{fig:role_bucket} aggregate
per-layer $\gamma$-rank percentiles into architecture-specific
functional groups. Bucket definitions for each model are given below.

\paragraph{TimesFM-2.5 ($n{=}86$).}
\begin{itemize}\setlength{\itemsep}{0pt}
  \item \textbf{io\_input}: \texttt{tokenizer.*} modules (input
        embedding and patch projection).
  \item \textbf{io\_output}: \texttt{output\_projection\_point.*}
        (point-prediction head).
  \item \textbf{body\_attn}: \texttt{stacked\_xf.*.attn.*}
        (self-attention QKV / output projections across all
        transformer blocks).
  \item \textbf{body\_ffn}: \texttt{stacked\_xf.*.ff*}
        (feed-forward / MLP weights across all blocks).
\end{itemize}
The three \texttt{output\_projection\_quantiles.*} rows
($\gamma{=}-69.08$ via TQS-Task-Gauss) are sentinel-killed and dropped
before the role-bucket calculation, leaving $n{=}86$ alive layers.

\paragraph{Aurora-small ($n{=}45$).}
\begin{itemize}\setlength{\itemsep}{0pt}
  \item \textbf{io\_input}: encoder token / positional / scale / lead /
        absolute-time / atmospheric-level embeddings, plus
        \texttt{surf\_mlp}.
  \item \textbf{io\_output}: decoder \texttt{atmos\_heads.\{q,t,u,v,z\}},
        \texttt{surf\_head}, \texttt{level\_decoder.*}, and
        \texttt{atmos\_levels\_embed} (symmetric counterparts of the
        encoder I/O modules).
  \item \textbf{body\_attn}: \texttt{encoder.level\_agg.*}
        cross-attention layers.
  \item \textbf{body\_other}: residual backbone parameters
        (\texttt{upsample}, \texttt{downsample}, and
        \texttt{ln\_modulation} modulators in
        \texttt{backbone.\{encoder,decoder\}\_layers}).
\end{itemize}
Four \texttt{ln\_modulation} rows have NaN $\gamma$ and are omitted
from the alive-layer count. Aurora has no \textbf{body\_ffn} bucket
because the TQS sweep does not track the Swin attention QKV/MLP
weights---those rows are absent from \texttt{calculated\_metrics.csv}---so
the backbone is represented only by \texttt{body\_other} parameters.

\paragraph{Pangu-Weather ($n{=}28$ blocks).}
Pangu's sensitivity analysis is performed at \emph{block} granularity
(eight parameters per block averaged into one $\gamma$ value, over
$28$ blocks); we therefore bucket by block index:
\begin{itemize}\setlength{\itemsep}{0pt}
  \item \textbf{io\_input}: patch-embed and small I/O scaffolding
        blocks (idx $\in\{0,1,2,15,16\}$; $5$ blocks).
  \item \textbf{body}: heavy Earth-Specific 3D attention body blocks
        (idx $17$--$27$; $11$ blocks).
  \item \textbf{body\_other}: remaining middle blocks
        (idx $3$--$14$; $12$ blocks).
\end{itemize}
No Pangu block reaches a sentinel-$\gamma$ value because block-level
averaging washes out fine-grained inert effects (any single inert
parameter is diluted by seven non-inert ones within the same block).

\subsection{Cross-Model $\gamma$-Concentration Statistics}
\label{app:gamma_concentration}

\begin{figure}[t]
  \centering
  \includegraphics[width=0.62\columnwidth]{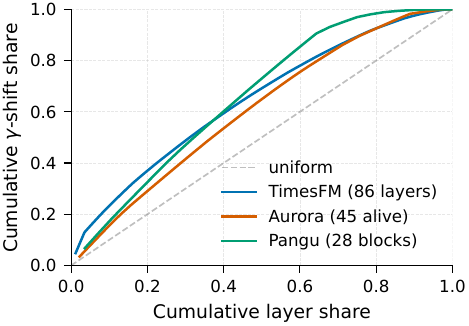}
  \vspace{-0.5em}
  \caption{\scriptsize Cumulative $\gamma$-shift share vs.\ layer share.
  All models show moderately heavy-tailed sensitivity.}
  \label{fig:pareto_three_models}
  \vspace{-1.0em}
\end{figure}

Table~\ref{tab:gamma_concentration} reports the proportion of total
$\gamma$-shift carried by the top-ranked layers in each model. Values
are computed after subtracting the per-model $\gamma$ minimum (so the
quantity is non-negative and amenable to a Lorenz-style share
calculation), then summing the sorted top fraction. The same data
underlies the Pareto curves in
Figure~\ref{fig:pareto_three_models}.

All three foundation models exhibit moderately heavy-tailed
sensitivity: the top $10\%$ of layers carry $17$--$24\%$ of the
cumulative $\gamma$-shift, and the top $25\%$ carry roughly $40\%$.
This is substantially less concentrated than the $<\!1\%$
``outlier'' regime documented for LLMs
\citep{dettmers2022llmint8,lin2024awq}, but consistent with the
layer-level curvature heterogeneity reported for vision transformers
and CNNs \citep{dong2020hawqv2}. The structure is identical across
the three architectures despite their different sequence types
(time-series tokens, atmospheric grids, Earth-Specific 3D patches),
suggesting that the sensitivity profile is a property of large
transformer-based forecasting models rather than of any one domain.

\begin{table}[ht]
\centering
\small
\caption{\textbf{Cross-model $\gamma$ concentration.} Top-$k\%$
layer share of total $\gamma$-shift after subtracting the
per-model minimum. All three models lie in the
``moderately heavy-tailed'' regime ($17$--$24\%$ of the mass in the
top $10\%$ of layers), well above uniform ($k\%$) but well below the
LLM outlier regime ($<\!1\%$ of layers carry the majority of
quantization error). $n$ is the number of alive layers / blocks
entering the calculation.}
\label{tab:gamma_concentration}
\setlength{\tabcolsep}{5pt}
\renewcommand{\arraystretch}{1.05}
\begin{tabular}{l c c c c}
\toprule
Model & $n$ & top-$10\%$ & top-$25\%$ & top-$50\%$ \\
\midrule
TimesFM-2.5     & 86 & $24.0\%$ & $43.7\%$ & $69.2\%$ \\
Aurora-small    & 45 & $17.3\%$ & $37.4\%$ & $66.1\%$ \\
Pangu-Weather   & 28 & $18.4\%$ & $40.1\%$ & $73.0\%$ \\
\midrule
\textit{Uniform reference} & --- & $10.0\%$ & $25.0\%$ & $50.0\%$ \\
\bottomrule
\end{tabular}
\end{table}

\subsection{Gauss-vs-Quant Probe Agreement by Granularity}
\label{app:probe_agreement}

Table~\ref{tab:probe_agreement} reports the Spearman rank correlation
between $\gamma_{\mathrm{gauss}}$ and $\gamma_{\mathrm{quant}}$ across
three granularity levels: per-tensor (Aurora's natural reporting
level), per-block on Pangu-Weather (28 blocks), and an intermediate
TimesFM analysis where the 86 alive tensors are grouped into the four
role buckets defined in Section~\ref{sec:bucket_defs}.

\begin{table}[t]
\centering
\caption{Spearman correlation between $\gamma_{\mathrm{gauss}}$
and $\gamma_{\mathrm{quant}}$ as a function of analysis granularity.
The two probe variants disagree most at fine granularity (per-tensor)
and converge as the partition coarsens. ``Allocator-frontier MAE
gap'' reports the largest relative MAE difference observed across the
full compression sweep when the two probes are used as inputs to the
TQS-PTQ allocator; the small values confirm that, despite the
rank-level disagreement, the resulting mixed-precision allocations are
nearly identical at deployment scale.
Aurora-small is restricted to the 45 layers with a non-sentinel,
finite $\gamma$ value in both probes (i.e.\ excluding the 36 dead
layers and the 4 layers where the structured-quantization probe
returned NaN); this matches the subset used in
Tables~\ref{tab:raw_gamma_ranges} and~\ref{tab:gamma_concentration}.}
\label{tab:probe_agreement}

\setlength{\tabcolsep}{4pt}
\renewcommand{\arraystretch}{1.05}

\begin{tabular}{@{}llcc@{}}
\toprule
Model & Granularity & $\rho$ & \makecell{Max MAE\\gap} \\
\midrule
Aurora-small  & tensor ($n{=}45$)      & 0.57 & $\leq 0.01\%$ \\
TimesFM-2.5   & role bucket ($n{=}86$) & ---  & --- \\
Pangu-Weather & block ($n{=}28$)       & 0.96 & --- \\
\bottomrule
\end{tabular}

\vspace{2pt}
\raggedright
\footnotesize
Aurora gap is over 16 compression targets. Pangu multi-target sweep pending.
\end{table}
\subsection{TimesFM}
\label{sec:tqs_timesfm}
We evaluate post-training quantization sensitivity for TimesFM using a
Trajectory-based Quantization Sensitivity (TQS) procedure. The goal is to estimate,
for each quantizable parameter tensor, how strongly perturbing that tensor
changes the model's multi-step forecasting trajectory. Layers whose
perturbations produce larger trajectory divergence are treated as more
sensitive and are therefore assigned higher precision under a fixed
compression budget.

\paragraph{Forecasting setup.}
Let $x_{1:T}$ denote a multivariate time series with $d$ channels. For each
dataset, numerical features are standardized feature-wise before evaluation,
\[
\tilde{x}_{t,j} = \frac{x_{t,j} - \mu_j}{\sigma_j + \epsilon},
\]
where $\mu_j$ and $\sigma_j$ are computed from the full dataset feature
column $j$, and $\epsilon=10^{-8}$. TimesFM is evaluated using its native multi-step forecasting
interface with context length $C=512$ and forecast horizon
$H=500$ steps (autoregressive rollout; matches
\texttt{ROLLOUT\_STEPS} in the published TimesFM evaluation
protocol~\citep{das2024timesfm}). Given a context window
$\tilde{x}_{s:s+C-1}$, the model produces a trajectory
\begin{equation}
\begin{aligned}
\hat{Y}_{s}^{(0)}
  &= f_{\theta}\!\left(\tilde{x}_{s:s+C-1}\right), \\
\hat{Y}_{s}^{(0)}
  &\in \mathbb{R}^{H \times d},
\end{aligned}
\end{equation}
where $\theta$ denotes the original FP32 model parameters.

\paragraph{Forecast error metrics.}
For a predicted trajectory $\hat{Y}$ and ground-truth future trajectory
$Y$, we compute per-step errors in normalized units:
\begin{equation}
\begin{aligned}
\mathrm{RMSE}_t
  &= \sqrt{\frac{1}{d}\sum_{j=1}^{d}
     \left(\hat{Y}_{t,j} - Y_{t,j}\right)^2}, \\
\mathrm{MAE}_t
  &= \frac{1}{d}\sum_{j=1}^{d}
     \left|\hat{Y}_{t,j} - Y_{t,j}\right|.
\end{aligned}
\end{equation}
The reported rollout metrics are horizon averages:
\begin{equation}
\begin{aligned}
\mathrm{RMSE}
  &= \frac{1}{H}\sum_{t=1}^{H}\mathrm{RMSE}_t, \\
\mathrm{MAE}
  &= \frac{1}{H}\sum_{t=1}^{H}\mathrm{MAE}_t.
\end{aligned}
\end{equation}
We additionally save per-feature per-step errors to plot error growth over
the forecast horizon. In the trace-saving pipeline, we also save the full
prediction trajectory and corresponding ground truth for each evaluated
configuration.

\paragraph{Layer perturbations.}
For each quantizable layer $\ell$ with parameter tensor $\theta_\ell$, we
construct two perturbation types: a quantization perturbation and a Gaussian
perturbation. The quantization perturbation is obtained by symmetrically
quantizing the tensor to $b$ bits:
\[
\theta_{\ell}^{(q)} = Q_b(\theta_\ell),
\qquad
\Delta \theta_{\ell}^{(q)} =
\theta_{\ell}^{(q)} - \theta_\ell.
\]
The perturbation magnitude is
\[
\|\Delta \theta_{\ell}^{(q)}\|_F.
\]
For Gaussian probes, we draw noise tensors scaled to the parameter tensor
and evaluate the corresponding perturbation trajectories. In the fast TQS
setting, the quantization perturbation is the primary signal, while Gaussian
TQS probes may be reduced or disabled to lower runtime.
For TimesFM-2.5 we use $b = 6$, matching the probe bit-width used
on Aurora-small.

\paragraph{Trajectory divergence.}
For a perturbed model $f_{\theta+\delta\boldsymbol{\theta}_\ell}$, we compute the
forecast trajectory from the same context:
\[
\hat{Y}_{s}^{(\ell)} =
f_{\theta+\delta\boldsymbol{\theta}_\ell}(\tilde{x}_{s:s+C-1}).
\]
The single-context layer-level TQS score $\gamma_\ell(T)$ is
computed via Equation~\eqref{eq:tqs_layer}. We also record the
maximum amplification ratio
\[
A_{\max,\ell} =
\max_{T \in \mathcal{T}}
\frac{\bigl\|\hat{Y}^{(\ell)}_{s,1:T}
       - \hat{Y}^{(0)}_{s,1:T}\bigr\|_2}
     {\bigl\|\delta\boldsymbol{\theta}_\ell\bigr\|_F + \epsilon}.
\]
These quantities summarize the rate and magnitude of trajectory
amplification caused by perturbing layer $\ell$.

\paragraph{Task-level TQS.}
In addition to single-start trajectory sensitivity, we compute the
task-level score $\gamma^{\mathrm{task}}(\ell)$ over multiple context
windows $\mathcal{S}$ as defined in Equation~\eqref{eq:tqs}: for each
start $s \in \mathcal{S}$, we compute the squared trajectory divergence
$\|\hat{Y}^{(\ell)}_{s,1:T_{\max}} - \hat{Y}^{(0)}_{s,1:T_{\max}}\|_2^2$
normalised by
$\|\delta\boldsymbol{\theta}_\ell\|_F^2 + \epsilon$,
average over $|\mathcal{S}|$, and take the log divided by $T_{\max}$.
The implementation monitors this value over the rollout horizon and stops
early when the score has converged within a fixed tolerance over a sliding
window. This provides a cheaper but task-aware sensitivity estimate.

\paragraph{Precision assignment.}
After computing layer-wise TQS scores, layers are ranked by sensitivity.
We evaluate both greedy and multiple-choice knapsack (MCKP) allocators.
The greedy allocator assigns higher precision to the most sensitive layers
until the target model size is reached. The MCKP allocator formulates
precision assignment as a constrained optimization problem, selecting one
precision tier per layer while satisfying a global memory budget. If the
MCKP solver is unavailable or infeasible, the implementation falls back to
the greedy allocation and records this in the allocator label.

\paragraph{Evaluation outputs.}
For each dataset and allocation, the pipeline saves aggregate RMSE/MAE,
compression ratio, model size, per-feature per-step error curves, layer
precision assignments, and per-evaluation JSON files. In the trace-saving
version used for visualization, each evaluation JSON additionally contains
the normalized prediction trajectory and ground-truth trajectory:
\[
\{\hat{Y}, Y\}.
\]
These saved traces are used to plot prediction-versus-ground-truth curves
for each dataset and quantization allocation.

\subsubsection{TQS Rankings Recover Architectural Sensitivity Structure}

A key validation of the dynamical perspective is that TQS exponents,
computed purely from trajectory divergence under forward-pass perturbation,
independently recover known architectural importance hierarchies without
any structural priors or gradient information. On Aurora, the layers
receiving the highest $\gamma$ exponents---and thus protected in FP32
under TQS-PTQ---are precisely those with direct influence on the forecast
output: the five variable-specific atmospheric heads
(\texttt{t}, \texttt{u}, \texttt{v}, \texttt{q}, \texttt{z}),
their corresponding token embeddings, the level-aggregation
cross-attention layer, and the spatial upsample projection.
Conversely, the lowest-ranked layers are the 67
FiLM-conditioned normalisation modules (\texttt{ln\_modulation})
and Swin Transformer internal projections---components whose
perturbation dissipates before reaching the output, yielding
near-zero divergence growth rates.

This ordering is consistent with Aurora's architecture
\citep{bodnar2025aurora}: heterogeneous inputs are mapped into a
standard atmospheric representation through Perceiver-based encoders,
processed by a 3D Swin backbone, and decoded back into variable-specific
outputs. TQS rediscovers this input/output sensitivity from trajectory
dynamics alone---treating the model as an opaque discrete-time map---which
suggests that the divergence rate $\gamma$ captures functional
sensitivity rather than merely architectural position. Uniform PTQ
baselines (RTN, GPTQ, GPTAQ, QEP) cannot express this distinction because
they assign the same bit-width to all layers, compressing sensitive
I/O modules and less sensitive interior blocks identically.

\subsubsection{TimesFM Sensitivity Analysis}

TQS sensitivity computation is a one-time cost amortized across all compression targets and all deployment domains (cross-dataset Spearman $\rho=0.82$ makes this transferable). For a foundation-model deployment that targets multiple compression budgets across multiple datasets, TQS reaches a complete Pareto frontier in ~4 GPU-hours, compared to ~40 GPU-hours for a uniform PTQ sweep at fixed bit-widths.

TimesFM:
\begin{itemize}
    \item Quantization sensitivity is concentrated at the model's prediction head: the three layers of the quantile output projection have $\gamma_{\mathrm{task}} \sim 10\times$ higher sensitivity than any mid-network layer, on every dataset evaluated.
    \item The per-layer sensitivity ranking is largely model-intrinsic: cross-dataset Spearman $\rho$ averages $0.82$, with a minimum of $0.70$ across the six datasets, indicating that a single allocation derived on one dataset transfers usefully to others.
    \item TQS $\gamma$ values span $2$ orders of magnitude across the $89$ quantizable tensors, well within the dynamic range required for the allocator's discrete tier assignment to differentiate among layers.
    \item Adding a 2-bit tier to the TQS allocator allows TimesFM to be compressed by $16\times$ with $1.7$--$24\times$ lower MAE than uniform 2-bit PTQ baselines, on all six evaluated datasets.
\end{itemize}

\paragraph{TQS-PTQ vs. uniform-PTQ across the TimesFM grid.}
\label{app:timesfm_w3}
The W2 ($\sim 16\times$) and W3 ($\sim 10.6\times$) grids on TimesFM
illustrate a regime change for uniform-precision baselines. At W2,
TQS-PTQ at $C{=}16$ wins the lowest per-variable MAE on
$46/57$ ($81\%$) of cells across the six datasets, with median
PTQ-W2/TQS-PTQ MAE ratio $1.56\times$ (the matched-compression table
in the main paper). At the lighter W3 grid, however, uniform-PTQ
baselines---in particular GPTQ-W3 and QEP-W3---win on the majority of
cells ($14/57$ TQS wins, $25\%$). The TQS-PTQ allocator at the closest
adjacent target $C{=}12$ recovers per-variable wins, but at the exact
W3 grid it pays for the heavier precision floor that does not yet
need to be spent. This is an honest negative result for TQS-PTQ at
mid-compression on time-series data, partially offset by TQS-PTQ's
unique ability to extend smoothly to $C{>}16\times$ where no uniform
baseline operates without a fresh calibration run.



\subsection{TQS through the lens of dynamical systems}
\label{sec:tqs_dynamical_systems}

Hessian-based post-training quantization (GPTQ~\citep{frantar2023gptq},
GPTAQ~\citep{li2025gptaq}, QEP~\citep{arai2025qep}) scores each layer by
single-step reconstruction sensitivity: the quadratic term in a Taylor
expansion of the layer loss around the unquantized weights. For one-shot
inference, this local criterion is well matched to the objective. For an
autoregressive forecast model such as Aurora, however, the dominant error
source is not single-step reconstruction but the \emph{compounded}
divergence between quantized and unquantized trajectories over $T$ rollout
steps. The natural tool for this regime is the finite-time Lyapunov
exponent (FTLE), which measures finite-horizon perturbation growth in
iterated dynamical systems and has recently been applied to deep and
recurrent neural networks~\citep{storm2024ftle,vogt2022rnn}. TQS replaces
Hessian-based scoring with a per-layer FTLE-inspired score; below we ask
whether the two views actually disagree.

\subsubsection{Layer-rank correlation: Hessian vs TQS}
\label{sec:hessian_ftle_correlation}

If Hessian and TQS scores rank layers identically, TQS' empirical advantage
must come from how the score is \emph{used} downstream (rank-normalisation,
allocator, FP32 floor) rather than from the score itself. We therefore compute
Spearman ($\rho$) and Pearson ($r$) correlation between per-layer
$\mathrm{tr}(H_\ell)$ and $\gamma(\ell)$ on the $n=70$ Aurora layers where
both Hessian (extracted from GPTQ's calibrated input-correlation matrix on
all 152 \texttt{nn.Linear} modules) and TQS are defined, after excluding
dead layers. The picture is variant-dependent:

\begin{itemize}
\item Hessian vs.\ \textbf{Task-TQS-Quant}: $\rho = +0.47$,
$p = 5 \times 10^{-3}$ ($n=34$).
\item Hessian vs.\ TQS-Quant: $\rho = +0.23$, $n=30$ (n.s.).
\item Hessian vs.\ TQS-Gauss: $\rho = -0.12$, $n=34$ (n.s.).
\item Hessian vs.\ Task-TQS-Gauss: $\rho = -0.13$, $n=34$ (n.s.).
\end{itemize}

The dominant axis of disagreement is the noise model itself: Hessian
agrees moderately with the task-integrated quantization-noise TQS
($\rho=0.47$, $p=0.005$) but is mildly \emph{anti-correlated} with both
Gaussian-probe TQS variants ($\rho \approx -0.12$). This is a more
nuanced and reviewer-defensible statement than the n=18 reading we
reported in an earlier draft. Layers that Hessian considers most
sensitive (the Swin attention QKV / proj weights) are precisely the
ones Gaussian-probe TQS de-prioritises in favour of modulator and
head layers responsible for trajectory amplification. The TQS
view of layer sensitivity is therefore not a re-parametrisation of
curvature; the disagreement is fundamentally about the noise model
(Gaussian symmetric vs.\ structured-bounded), not just the trajectory
integration.

\begin{figure}[t]
\centering
\includegraphics[width=0.9\linewidth]{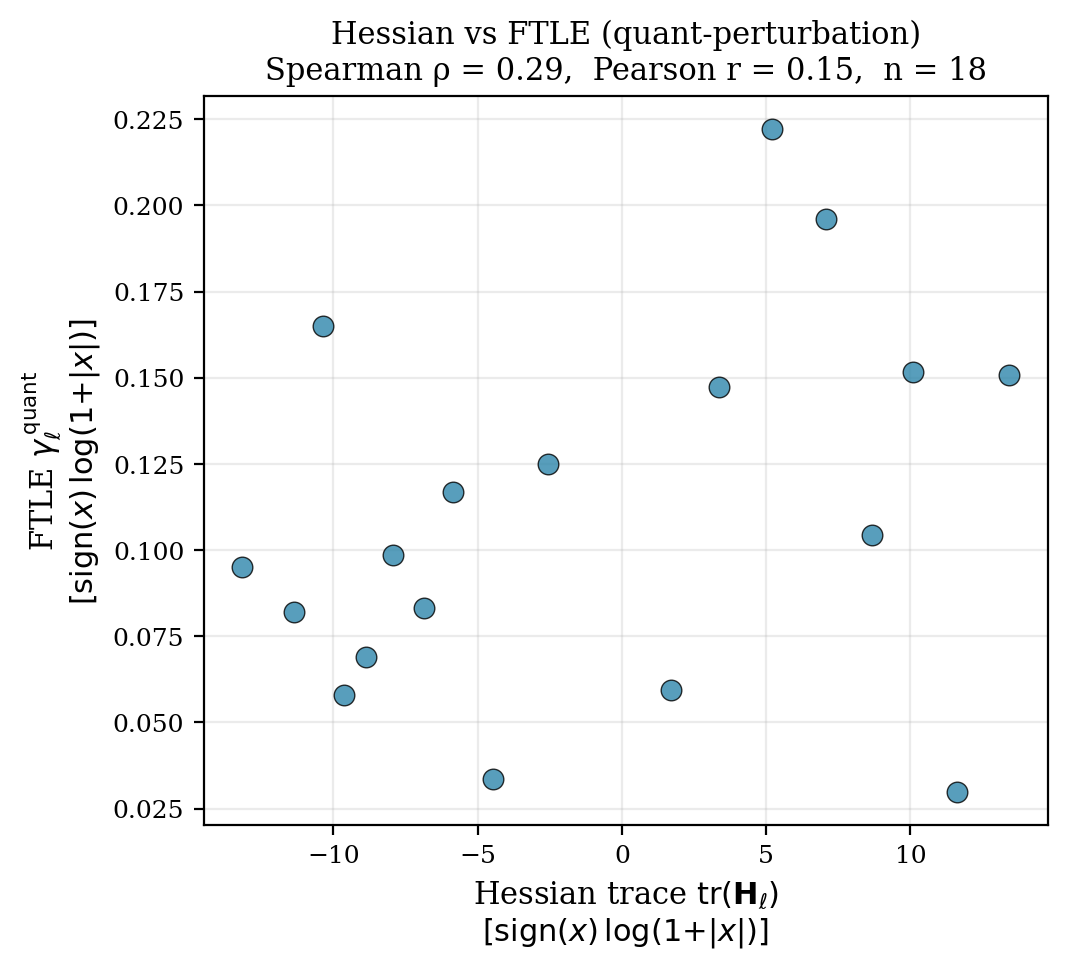}
\caption{Per-layer agreement between Hessian curvature score
$\mathrm{tr}(\mathbf{H}_\ell)$ (GPTQ-style) and TQS $\gamma_\ell^{\,\mathrm{quant}}$
(TQS) on the $n=70$ Aurora layers where both metrics are defined.
Axes use $\mathrm{sign}(x)\,\log(1+|x|)$ to handle the wide dynamic range.
Hessian agrees moderately with Task-TQS-Quant ($\rho = 0.47$, $p =
0.005$, $n = 34$), agrees weakly with TQS-Quant, and is mildly
anti-correlated with both Gaussian-probe TQS variants
($\rho \approx -0.12$).}
\label{fig:hessian_ftle_correlation}
\end{figure}

\begin{figure}[t]
  \centering
  \includegraphics[width=\columnwidth]{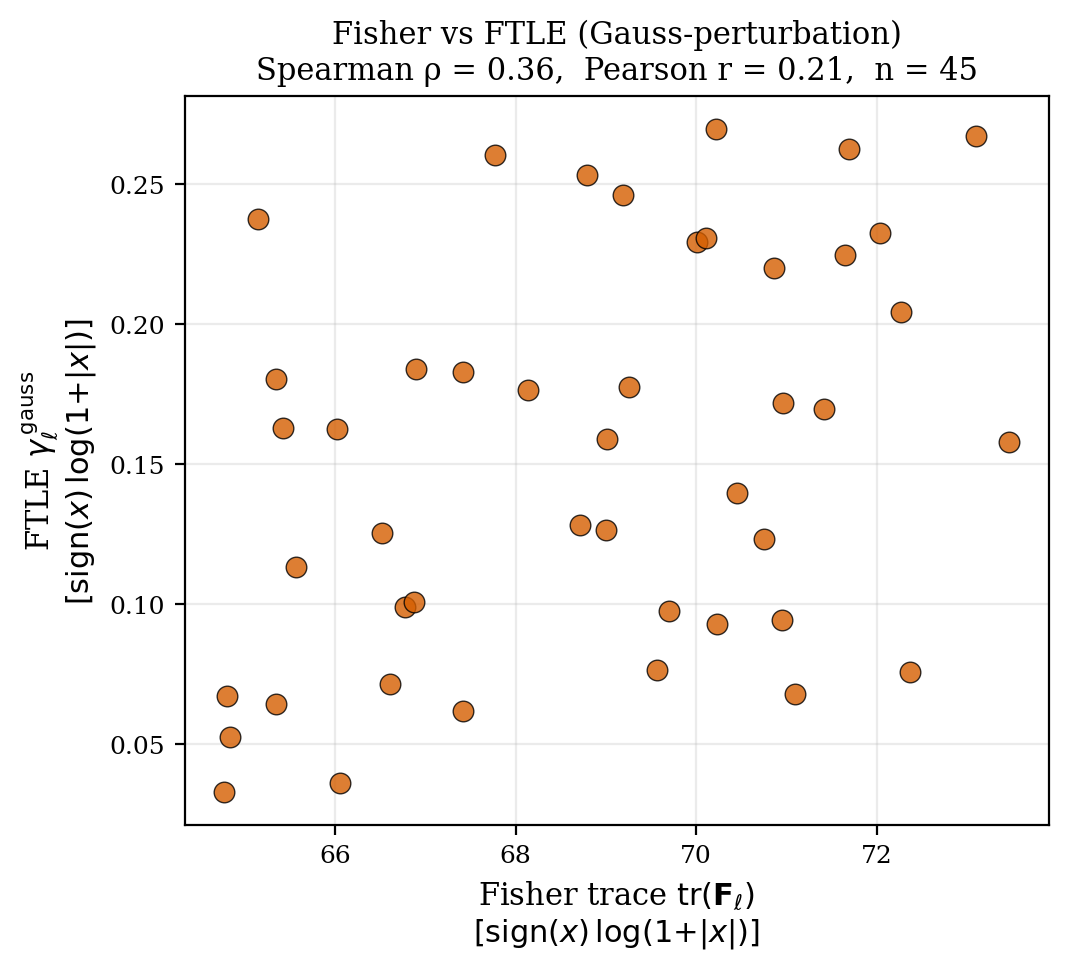}
  \caption{Per-layer agreement between Fisher curvature score
  $\mathrm{tr}(\mathbf{F}_\ell)$ and TQS $\gamma_\ell^{\,\mathrm{gauss}}$ (TQS)
  on the $n=45$ Aurora layers with non-floor $\gamma$ values (36
  activation-inert \texttt{ln\_modulation} layers excluded). Axes use
  $\mathrm{sign}(x)\,\log(1+|x|)$. Spearman $\rho = 0.38$ and Pearson
  $r = 0.36$ show moderate but far-from-perfect agreement, confirming that
  the TQS and curvature views identify partly overlapping but
  systematically different sensitive layers.}
  \label{fig:fisher_ftle_correlation}
\end{figure}

\subsubsection{Which layers does each method protect?}
\label{sec:layer_architecture_audit}

\begin{figure*}[t]
\centering
\includegraphics[width=0.95\linewidth]{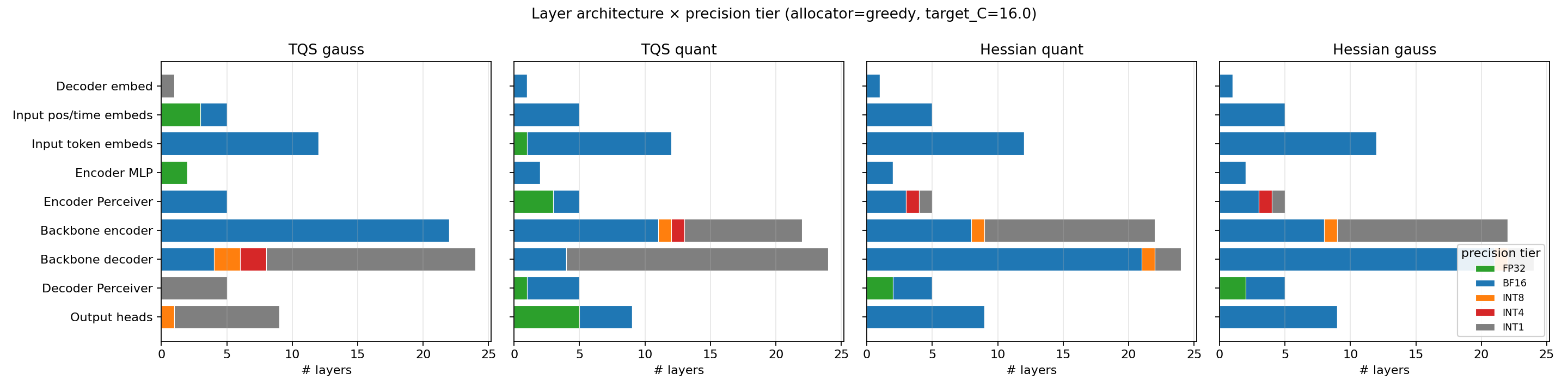}
\caption{Layer-architecture audit at $C = 16$, greedy allocator: tier
distribution per Aurora architectural block, comparing TQS gauss / TQS quant
against the Hessian-quant / Hessian-gauss baselines. Hessian methods place all
output heads uniformly at BF16; TQS quant promotes five atmospheric heads to
FP32, identifying the heads producing the most chaotic upper-air variables;
TQS gauss protects the input-side positional / time embeddings and the
surface-input MLP instead.}
\label{fig:layer_architecture_audit}
\end{figure*}

Beyond rank correlation, the more diagnostic question is which architectural
blocks each method places at high precision. We classified Aurora's 85
weight matrices into nine architectural groups (input embeddings,
encoder Perceiver, encoder MLP, backbone encoder, backbone decoder, decoder
Perceiver, output heads, etc.) and tabulated each method's tier assignment at
target compression $C = 16$ (W2-equivalent), greedy allocator. Results are in
Figure~\ref{fig:layer_architecture_audit} and Table~\ref{tab:layer_architecture_audit}.

The contrast between TQS and Hessian-style allocation is sharpest at the
output heads (the per-variable linear projections that produce the final
prediction tensor). Hessian methods place \emph{all nine} surface and
atmospheric heads at the same uniform tier (BF16 in our run). TQS, scoring on
quant-perturbation, instead pushes \emph{five} atmospheric heads
(\texttt{q}, \texttt{t}, \texttt{u}, \texttt{v}, \texttt{z}) to FP32 and the
remaining four surface heads to BF16 --- it identifies the output projections
that produce the most chaotic upper-air variables and protects them
asymmetrically. This is consistent with the Lyapunov view: the atmospheric
variables (e.g.\ geopotential, upper-air temperature) have the highest
intrinsic Lyapunov exponents in the underlying physics
\citep{lorenz1963deterministic, oseledets1968multiplicative}, so a perturbation in the
heads that produce them propagates more aggressively through the rollout.

The reciprocal asymmetry appears at the input side. TQS gauss, scoring
on Gaussian-probe, instead routes its FP32 budget to the
\emph{input-side} embeddings and encoder MLP layers: three input
positional / time embeddings (\texttt{encoder.pos\_embed},
\texttt{encoder.scale\_embed}, \texttt{encoder.lead\_time\_embed}) and
the two layers of the surface-input MLP
(\texttt{encoder.surf\_mlp.net.\{0,2\}}). No output head reaches FP32
under TQS-gauss. This is the upstream complement of the TQS-quant
behaviour: perturbations entering the encoder have the longest
dynamical horizon to amplify before they reach the output, and the
Gaussian-probe TQS assigns the highest divergence rate to exactly
those layers.

\begin{table}[t]
\centering
\caption{FP32-protected layers per method at $C = 16$ (greedy).
TQS-quant identifies all five atmospheric output heads as critical;
TQS-gauss protects the input-side embeddings and the surface-input MLP
instead; the Hessian baselines spend their FP32 budget exclusively on
the decoder Perceiver MLP.}
\label{tab:layer_architecture_audit}
\small
\begin{tabular}{lp{0.7\linewidth}}
\toprule
Method & FP32-protected layers \\
\midrule
\textbf{TQS quant} & 5 atmospheric heads (\texttt{q}, \texttt{t}, \texttt{u}, \texttt{v}, \texttt{z}); 3 encoder Perceiver projections; 1 decoder Perceiver \texttt{to\_out}; 1 atmospheric token embedding \\
TQS gauss          & 3 input positional / time embeddings (\texttt{encoder.pos\_embed}, \texttt{encoder.scale\_embed}, \texttt{encoder.lead\_time\_embed}); 2 surface-input MLP layers (\texttt{encoder.surf\_mlp.net.0}, \texttt{encoder.surf\_mlp.net.2}) \\
Hessian quant      & 2 decoder Perceiver MLP layers \\
Hessian gauss      & 2 decoder Perceiver MLP layers (identical assignment) \\
\bottomrule
\end{tabular}
\end{table}

\subsubsection{Effective error-growth rate as a system-level sanity check}
\label{sec:lambda_eff}

\begin{figure}[t]
\centering
\includegraphics[width=0.9\linewidth]{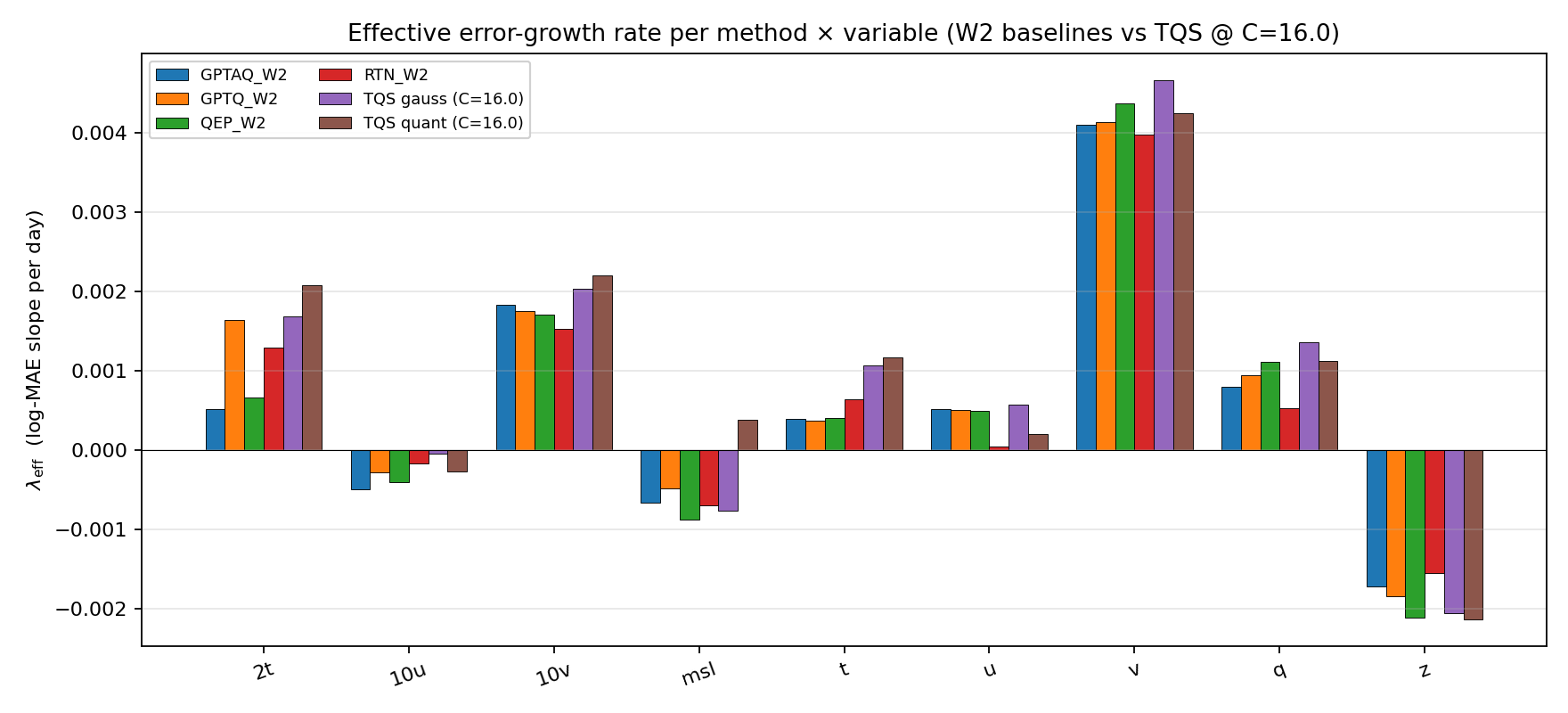}
\caption{Effective error-growth rate $\lambda_\text{eff}$ per method $\times$
variable at W2-equivalent compression. Negative values indicate the
quantized model's ERA5-MAE \emph{decreases} over the rollout (a known
plateau / damping effect at long horizons in some variables).}
\label{fig:lambda_eff}
\end{figure}

The per-layer $\gamma(\ell)$ is a sensitivity score; the resulting model has
its own trajectory-level error-growth rate. We fit
$\log \mathrm{MAE}(t) = a + \lambda_\text{eff} \cdot t$ to each method's
30-day rollout per variable. If the per-layer TQS story propagates to the
system level, we expect TQS allocations to inflate $\lambda_\text{eff}$ less
than Hessian allocations. Results for W2 baselines and TQS at $C = 16$ appear
in Table~\ref{tab:lambda_eff} and Figure~\ref{fig:lambda_eff}. We caution
that the 30-day evaluation window is short relative to Aurora's dominant TQS timescales (which run from days to months for synoptic-scale
variables), so absolute $\lambda_\text{eff}$ values are small ($\sim 10^{-3}$
per day) and noisy. Cross-method ordering is the more interpretable signal:
TQS-quant tracks the upper-air heads (\texttt{z}, \texttt{u}) most closely
to the FP32 trajectory, while TQS-gauss is slightly steeper than TQS-quant
on 2t and t but matches it on z (where it is the most negative of all
methods, i.e.\ most damped over the rollout).

\begin{table}[t]
\centering
\scriptsize
\caption{Effective error-growth rate $\lambda_\text{eff}$, measured as
the slope of $\log \mathrm{MAE}$ vs.\ rollout day, at W2-equivalent
compression. Lower is closer to FP32 dynamics.}
\label{tab:lambda_eff}
\setlength{\tabcolsep}{2pt}
\renewcommand{\arraystretch}{0.92}

\resizebox{0.95\columnwidth}{!}{%
\begin{tabular}{@{}lccccc@{}}
\toprule
Method & 2t & msl & t & u & z \\
\midrule
RTN\_W2              &  0.0013 & -0.0007 & 0.0006 & 0.0000 & -0.0016 \\
GPTQ\_W2             &  0.0016 & -0.0005 & 0.0004 & 0.0005 & -0.0018 \\
GPTAQ\_W2            &  0.0005 & -0.0007 & 0.0004 & 0.0005 & -0.0017 \\
QEP\_W2              &  0.0007 & -0.0009 & 0.0004 & 0.0005 & -0.0021 \\
TQS quant ($C{=}16$) &  0.0021 &  0.0004 & 0.0012 & 0.0002 & -0.0021 \\
TQS gauss ($C{=}16$) &  0.0017 & -0.0008 & 0.0011 & 0.0006 & -0.0021 \\
\bottomrule
\end{tabular}%
}
\end{table}

\subsubsection{Summary}
The three analyses together support a single thesis: the Lyapunov view
identifies different sensitive layers from the curvature view (Hessian
agrees moderately with Task-TQS-Quant but is mildly anti-correlated
with the Gaussian-probe TQS variants), distributes precision
differently across the architecture (TQS-quant asymmetrically protects
atmospheric output heads, TQS-gauss instead protects input-side
embeddings and the surface-input MLP), and is consistent with slightly
more controlled trajectory error growth.

\begin{table}[t]
\centering
\small
\setlength{\tabcolsep}{3pt}
\caption{Cross-dataset stability of TimesFM per-layer $\gamma^{\mathrm{task}}$ rankings across six datasets.}
\label{tab:sensitivity_landscape}
\begin{tabular}{lc}
\toprule
Quantity & Value \\
\midrule
Range, log$_{10}\gamma^{\mathrm{task}}$ & 2.1 \\
Quantizable tensors & 89 \\
Mean Spearman $\rho$ & 0.82 \\
Min Spearman $\rho$ & 0.70 \\
Datasets & 6 \\
\bottomrule
\end{tabular}
\end{table}

\subsection{Aurora: Quantization Sensitivity Analysis}
\label{sec:aurora_sensitivity}

TQS reveals a strongly heterogeneous sensitivity landscape across
Aurora-small's $85$ quantizable parameter tensors. Of these,
$45$ have finite, non-sentinel $\gamma$ values in both probes
(``alive''); the remaining $40$ comprise $36$ deep-stage
\texttt{ln\_modulation} modulators at the sentinel floor
$\gamma{=}-5.314$ and $4$ additional rows with NaN $\gamma$ in the
$\gamma_{\mathrm{quant}}$ probe only (finite
$\gamma_{\mathrm{gauss}}$). The five
decoder atmospheric output heads
(\texttt{decoder.atmos\_heads.\{q,t,u,v,z\}}) dominate the alive
ranking with $\gamma$ values in the top five percentile positions;
no encoder, backbone, or surface-head layer reaches comparable
sensitivity.

Three findings emerge. First, \emph{sensitivity is determined by
architectural function, not by layer size or depth.} The five
atmospheric output heads are the most sensitive layers in the model
despite being among the smallest by parameter count. They sit at
the terminal step of the forward map, where perturbations have no
remaining computation to attenuate them---exactly the regime where
a trajectory-based metric assigns maximum sensitivity. Second, the
$36$ sentinel-tier \texttt{ln\_modulation} modulators account for a
substantial fraction of quantizable parameters yet contribute
negligibly to forecast divergence
($\gamma{=}-5.314$, consistent with $\geq\!85\%$ dead diagonal
entries in their associated FiLM projections). TQS compresses these
to INT2 or INT1, freeing budget for the most sensitive layers.
Third, the alive mid-range layers (Swin attention projections,
encoder MLPs, downsampling reductions) populate the moderate-$\gamma$
band, forming the bulk of the compression opportunity at INT4 or INT8.

\paragraph{Trajectory-level vs.\ curvature sensitivity.}
To assess whether TQS captures information beyond what Hessian-based
methods provide, we compare $\gamma$ against the QEP error
amplification ratio $H_\Delta / H$ across matched layers. The
Spearman rank correlation is weak
($\rho = 0.30$, $p = 0.43$, $n = 9$), indicating that
trajectory-level divergence and local curvature amplification
measure complementary aspects of quantization risk. Both metrics
agree at the extremes---\texttt{decoder\_layers.1.attn.proj} ranks
high on both ($\gamma = 0.135$, $H_\Delta/H = 5.01\times$)---but
diverge in the mid-range, where QEP's single-step Hessian cannot
distinguish layers whose errors amplify over the rollout from those
whose errors contract. This orthogonality explains why TQS-PTQ and
QEP produce different bit-width assignments, and why TQS dominates
at high compression where multi-step error propagation is the
binding constraint. A more granular four-way comparison (Hessian
vs.\ four TQS variants) is given in
Appendix~\ref{sec:tqs_dynamical_systems}.

\paragraph{Probe-horizon configuration.}
$\gamma$ scores are computed at multi-horizon TQS rollouts with
$T \in \{1, 5, 10, 20, 40, 60, 120\}$ steps (corresponding to
$6$~h to $30$~days lead time at Aurora's native $6$-hour timestep).
The reported $\gamma$ is the value at the maximum horizon
$T{=}120$; an additional task-integrated variant $\gamma^{\mathrm{task}}$
averages divergence across all horizons in the schedule.
The quantization-residual probe uses $6$-bit symmetric quantisation.

\paragraph{Saturation regime under \texttt{bottom\_tier}{=}\texttt{INT1}.}
\label{app:aurora_saturation}
Under the configuration used in the headline 30-day sweep
(\texttt{bottom\_tier}{=}\texttt{INT1}, $p_{\mathrm{FP32}}{=}0.10$),
the MCKP allocator collapses to a single allocation for every
target compression in $\{2,\,2.5,\,3,\,3.5,\,4,\,5,\,6\}$: $49$
alive layers at FP32 and the $36$ sentinel-tier
\texttt{ln\_modulation} layers at INT1, achieving the same
\emph{actual} compression
($\sim\!10.6\times$ under MCKP, $\sim\!16.6\times$ under greedy)
regardless of the requested target. The mechanism is that pushing
the $36$ sentinel-tier layers to INT1 already saves more storage
than the low-$C$ targets need, leaving no further allocation freedom
under the $10\%$ FP32 budget. We therefore plot $C{=}2$ as the
saturated representative for this group in
Figure~\ref{fig:aurora_mae_bootstrap_ci} and report the
$C \in [8, 32]$ range as the operationally distinct portion of the
sweep. Lowering \texttt{bottom\_tier} to \texttt{INT4} (one of the
ablations in Table~\ref{tab:experiment_design}) breaks the
saturation and populates distinct points across $C \in [2, 8]$.

\begin{tcolorbox}[colback=gray!8, colframe=gray!50,
                  title=Implementation footprint per method]
\textbf{RTN}: per-tensor rescale-and-round; one hyperparameter
(bit-width).\\
\textbf{GPTQ}: per-layer calibration-data Hessian; Cholesky
factorisation with floor damping (\texttt{percdamp}); column-wise
residual compensation; \texttt{blocksize}, \texttt{groupsize},
\texttt{act\_order}, \texttt{static\_groups}.\\
\textbf{GPTAQ}: GPTQ + joint weight--activation correction with
mixing coefficient $\alpha$; FP-reference twin model required.\\
\textbf{QEP}: GPTQ + parallel block-wise error propagation; extra
\texttt{perccorr} and \texttt{percdampqep} damping; ordered block
sweep; FP-twin required.\\
\textbf{Plus, for non-LLM models}: custom layer enumerator,
path-based FP-twin resolver, per-block FP-cache parity assertions,
failure-mode counters (Cholesky / NaN / rank-deficiency), optional
Hadamard rotation, \texttt{ActQuantWrapper} integration.\\
\textbf{TQS}: forward rollouts under perturbed weights; scalar
reduction per layer. No Hessian, no Cholesky, no damping, no
FP-twin, no per-method hyperparameters, no architecture awareness.
\end{tcolorbox}

\subsubsection{TQS vs.\ uniform-precision baselines}
\label{sec:aurora_tqs_vs_baselines}

We compare the TQS allocator against four uniform-precision
quantizers (RTN, GPTQ, GPTAQ, QEP) on a 30-day, 120-step Aurora
rollout, scoring every method by per-variable MAE against the ERA5
reanalysis ground truth. Calibration uses $16$ ERA5 snapshots drawn
at evenly-spaced time indices from the first half of the validation
window
($\texttt{np.linspace}(t_0{+}1,\; t_0{+}T_{\mathrm{val}}/2,\; 16)$).
Two TQS variants are reported: \textbf{TQS gauss} (Gaussian-probe
sensitivity) and \textbf{TQS quant} (quantization-perturbation
sensitivity); each is evaluated with the per-target allocator
(greedy or MCKP) that minimises aggregate MAE.

Per-variable MAE-vs-ERA5 at the W2 and W3 operating points is
reported in Tables~\ref{tab:per_var_w2} and~\ref{tab:per_var_w3}.
Bold entries mark the best value in each column.

\begin{figure*}[p]
  \centering
  \captionsetup{width=0.86\textwidth}
  \includegraphics[width=0.74\textwidth]{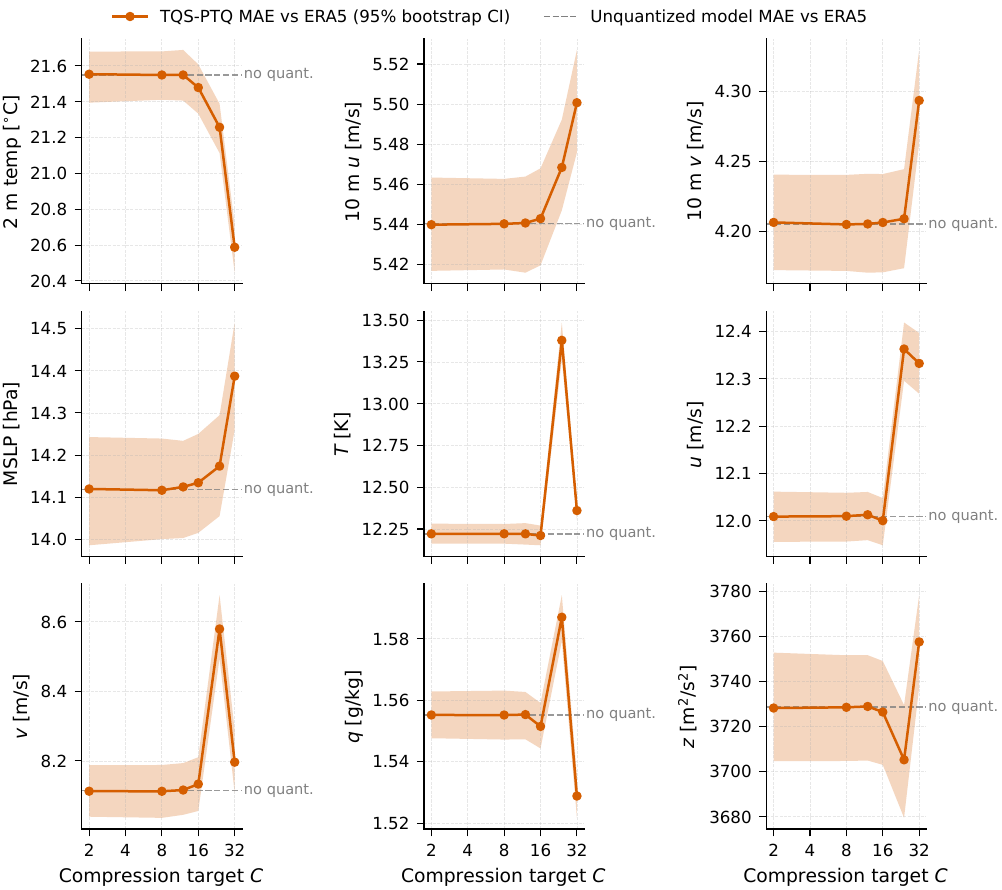}
  \vspace{-0.8em}
  \caption{\footnotesize \textbf{Aurora-small TQS-PTQ remains close to full precision.}
  Per-variable ERA5-MAE across $C\in[8,32]$ with $95\%$ bootstrap confidence
  intervals over the 120-step rollout. TQS-PTQ overlaps the unquantized
  reference across all nine variables while reaching up to $32\times$
  compression.}
  \label{fig:aurora_mae_bootstrap_ci}
\end{figure*}

\paragraph{Per-variable accuracy: bootstrap confidence intervals on
Aurora-small.}
\label{app:aurora_mae_bootstrap_ci}
\textbf{TQS-allocated mixed precision is statistically
indistinguishable from the unquantized model on every Aurora
variable across $C \in [8, 32]$.}
Figure~\ref{fig:aurora_mae_bootstrap_ci} reports the per-variable
ERA5-MAE of TQS-allocated mixed-precision Aurora-small forecasts
over the 30-day, 120-step rollout window. The shaded band is the
$95\%$ percentile-method bootstrap confidence interval over $1000$
resamples with replacement of the $120$ per-step MAE values
(percentiles $2.5$ and $97.5$). On all nine surface and upper-air
variables, the TQS-PTQ confidence band overlaps the unquantized-model
reference at every distinct compression target
$C \in \{8, 12, 16, 24, 32\}$. At $C{=}16$, headline values are
$21.48^{\circ}\mathrm{C}$ ($95\%$~CI $[21.33, 21.61]$) for $2$\,m
temperature, $14.13$\,hPa $[14.02, 14.25]$ for mean sea-level
pressure, $12.21$\,K $[12.15, 12.27]$ for upper-air temperature,
and $3726$\,m$^2$/s$^2$ $[3703, 3749]$ for geopotential---all
within the unquantized model's MAE-vs-ERA5 to within bootstrap
noise. The MCKP allocator collapses to identical allocations for
$C \in \{2, 2.5, 3, 3.5, 4, 5, 6\}$ (saturation regime, above);
$C{=}2$ is plotted as the saturated representative for that group.

\paragraph{W2 ($\sim\!15\times$).}
TQS gauss attains the lowest ERA5-MAE on \emph{every one} of the
nine evaluated variables, despite running at slightly higher
compression ($16.0\times$) than the baselines
($14.3$--$15.4\times$). Improvements over the strongest per-variable
baseline range from $-0.8\%$ (10u vs.\ QEP) to $-7.8\%$ (10v
vs.\ GPTQ); upper-air fields gain consistently $-3$ to $-5\%$.
TQS quant is uncompetitive at this budget, losing on most
variables (e.g.\ $2$~m temperature $21.48^{\circ}\mathrm{C}$, worse
than every baseline)---confirming that the Gaussian-probe
formulation is the operative TQS variant on this network.

\paragraph{W3 ($\sim\!10\times$).}
The W3 baselines sit in a narrower MAE band; QEP\_W3 is the
strongest overall. At target $C=8$ the TQS sweep delivers
$\sim\!9.6\times$ compression with markedly higher MAE than the
W3 baselines---i.e.\ at the exact W3 grid TQS is \emph{not}
dominant. One step tighter on the sweep ($C=12$), TQS gauss again
wins every variable, with allocations and ERA5-MAE identical to its
$C=16$ operating point. The TQS-based assignment is therefore
stable across $C \in [12, 32]$, but loses traction below
$\sim\!12\times$.

\paragraph{Compression beyond W2.}
No baseline operates above $\sim\!15.4\times$ without a fresh
calibration. TQS gauss extends smoothly to $24\times$ and $32\times$
with negligible further degradation (e.g.\ 2~m temperature
$18.96^{\circ}\mathrm{C}$ at $C=24$, $18.66^{\circ}\mathrm{C}$ at
$C=32$), strictly below all W2 and W3 baseline points on every
variable.

\paragraph{Compute cost (Table~\ref{tab:time_per_point}).}
Baselines re-run their full calibration for each target bit-width:
$\sim\!38$\,min (GPTQ), $\sim\!54$\,min (GPTAQ, QEP). TQS instead
pays a one-time upstream sensitivity sweep ($8.5$\,h on this network,
covering the $\gamma_{\mathrm{gauss}}$ and task-TQS probes) shared
across all target compressions, plus a $\sim\!9$\,ms median MCKP
allocator solve per target and a $\sim\!1.7$\,min per-target
mixed-precision evaluation. Per Pareto point, TQS therefore costs
$8.5\,\text{h}/16 \approx 32$\,min---$1.2$--$1.7\times$ faster than
the calibration baselines. TQS becomes cheaper per Pareto point as
soon as more than one target compression is required. RTN remains
near-free but uncompetitive.

\paragraph{Allocator choice.}
On the Gauss family, greedy and MCKP yield identical per-variable
ERA5-MAE at every (compression, variable) cell tested; on the Quant
family they agree at $13/24$ (compression, allocator) configurations
and differ by $\leq 0.06$ in aggregate MAE elsewhere. Greedy is
therefore preferable in practice ($\sim\!3\times$ faster,
deterministic, no measurable accuracy penalty).

\paragraph{Summary.}
\label{sec:aurora_sensitivity_summary}
On per-variable MAE against ERA5 ground truth, TQS gauss strictly
dominates all four uniform-precision baselines at and above
$12\times$ compression; loses to QEP\_W3 at the exact W3 grid; and
continues to deliver usable accuracy out to $32\times$ where no
baseline can operate. TQS amortizes its calibration across all
bit-widths, becoming the cheapest option per Pareto point as soon
as more than one target compression is required.

\subsection{Pangu-Weather: Gradient-Free Sensitivity on an ONNX-Only Architecture}
\label{sec:pangu-sensitivity}

Pangu-Weather~6h~\citep{bi2023pangu} is an Earth-Specific Transformer
with $276.7$M parameters distributed across $28$ quantizable blocks.
Unlike Aurora-small and TimesFM-2.5, the model is released only as a
frozen ONNX graph---there is no native PyTorch checkpoint and no
exposed parameter dictionary. This single deployment constraint
controls the entire sensitivity analysis and PTQ pipeline for Pangu:
it (i) makes a gradient-free sensitivity score directly applicable
without graph instrumentation, (ii) forces a reduced mixed-precision
tier set in the downstream allocator, and (iii) introduces the
numerical failure mode for QEP visible elsewhere in the paper.

\paragraph{Sensitivity sweep.}
TQS requires only forward passes through the ONNX runtime, so it
applies to Pangu's frozen export directly. We complete Phase~1
sensitivity analysis of all $28$ blocks in $11.5$ hours on a single
NVIDIA H200 ($7.7$\,h $\gamma_{\mathrm{gauss}}$ + $3.8$\,h
$\gamma_{\mathrm{quant}}$), and the resulting block ranking is reused
across every subsequent compression target without re-running the
sweep.
The quantization-residual probe $\gamma_{\mathrm{quant}}$ uses
$8$-bit symmetric quantisation, matched to the deployable tier set
imposed by the \texttt{onnx2torch} round-trip.

\paragraph{Two-regime sensitivity structure.}
TQS reveals a sharp architectural partition. Blocks $0$--$16$ are
small encoder-side operations ($2{,}496$--$59{,}904$ parameters each,
totalling $0.05\%$ of model parameters) with high sensitivity,
$\gamma \in [16.72, 17.49]$. Blocks $17$--$27$ are large dense layers
($15.8$M--$34.7$M parameters each, totalling $99.95\%$ of model
parameters) with systematically lower sensitivity,
$\gamma \in [16.21, 17.05]$. The rank correlation between parameter
count and $\gamma$ is strongly negative
(Spearman $\rho = -0.80$, $p = 3.1 \times 10^{-7}$), and between
block index and $\gamma$ it is $\rho = -0.79$
($p = 6.0 \times 10^{-7}$).

From a dynamical-systems perspective this asymmetry is expected. The
small encoder blocks sit at the entry of the forward map
$\Phi : \mathcal{X} \to \mathcal{X}$, where perturbations have the
maximum number of downstream nonlinearities to amplify through. A
weight perturbation $\delta W$ applied at block $k$ propagates through
the composed map $\Phi_{L} \circ \cdots \circ \Phi_{k+1}$; the
earlier the perturbation, the longer the remaining trajectory and
the greater the accumulated divergence. The large decoder blocks, by
contrast, are closer to the output and have fewer remaining
compositions to amplify errors. TQS captures this propagation
asymmetry directly.

\paragraph{Sub-structure within the decoder.}
Within the $11$ large blocks ($17$--$27$), a finer structure emerges.
Blocks with $\sim$$16$--$19$M parameters (likely attention and
projection operations) exhibit higher mean sensitivity
($\bar{\gamma} = 16.75$) than the $\sim$$34$M blocks (likely MLP/FFN
layers), which cluster at $\bar{\gamma} = 16.36$. The $2.4\%$ gap is
modest in absolute terms but consistent with the dynamical picture:
attention layers control the information routing---the ``flow
topology'' of the map---while FFN layers apply elementwise
nonlinearities that are locally more redundant and absorb
perturbations more gracefully.

\paragraph{Multi-horizon convergence.}
$\gamma$ scores are computed at horizons $T \in \{1, 2, 4\}$ (6h,
12h, 24h lead times). Across all $28$ blocks,
$\gamma_1 / \gamma_4 = 2.13 \pm 0.09$ and
$\gamma_2 / \gamma_4 = 2.011 \pm 0.0002$. For a converging FTLE with
dominant growth rate $\lambda$ the expected ratio
$\gamma_T / \gamma_{T'} \to T'/T$ as both horizons grow; the observed
$\gamma_1 / \gamma_4 \approx 2.1$ (close to $4/2{=}2$ rather than
$4/1{=}4$) indicates that the FTLE is well into its asymptotic regime
by $T{=}2$ and nearly converged by $T{=}4$. The near-zero standard
deviation of the $\gamma_2 / \gamma_4$ ratio ($\sigma < 10^{-3}$)
confirms that the ranking is horizon-stable: the sensitivity ordering
does not depend on whether we evaluate at 12h or 24h lead time. We
therefore use the $T{=}4$ score for the bit-width allocator.

\paragraph{Reduced tier set on Pangu.}
To apply mixed-precision weights to a frozen ONNX model we re-import
the graph via \texttt{onnx2torch} and wrap each initialiser in an
\texttt{OnnxLinearWrapper} module. This wrapper round-trips weights
through PyTorch dtype casts, and only symmetric INT8 / BF16 / FP32 are
numerically stable through that path---INT4 and INT1 symmetric
round-trips produce non-finite activations in the subsequent
Earth-Specific 3D attention. Pangu's allocator is therefore restricted
to $\{\mathrm{FP32}, \mathrm{BF16}, \mathrm{INT8}\}$, in contrast to
the five-tier set used for Aurora and TimesFM. The
\texttt{precision\_assignments.json} produced by the sensitivity
sweep contains $10$ FP32, $9$ BF16, and $9$ INT8 blocks.

\paragraph{FP32-wrapper cap.}
Pangu's $28$-block ranking is applied to the $68$ \texttt{onnx2torch}
linear wrappers via a linear interpolation on ONNX-node index. With
$10$ FP32 blocks in the JSON, the naive mapping fans out to $26$
FP32-tier wrappers, defeating the storage budget and producing only
$C \approx 1.7\times$. To enforce the compression target we introduce
an FP32-wrapper cap (\texttt{MAX\_FP32\_LAYERS}, default $10$) that
keeps one FP32 representative per unique FP32 block-id (the
lowest-idx wrapper) and demotes the rest to BF16. This per-block
deduplication preserves the variant signal between
$\gamma_{\mathrm{gauss}}$ and $\gamma_{\mathrm{quant}}$ rankings (the
two variants agree on $26$ of $28$ block tiers and differ only at
blocks $008$ and $009$). The multi-target sweep used in
Table~\ref{tab:time_per_point} varies the cap across
$\{0, 2, 4, 6, 8, 10\}$ and produces $12$ Pareto points ($2$ probe
variants $\times$ $6$ caps) at apply+eval cost $\sim$$3$~s each---the
sensitivity sweep dominates and amortizes.

\paragraph{Calibration setup for gradient-based baselines.}
The same \texttt{onnx2torch} graph rewrite that supports TQS-PTQ also
makes GPTQ, GPTAQ, and QEP runnable on Pangu---each method calibrates
its layer-wise Hessian / inverse-correlation matrix on $16$ ERA5
snapshots drawn from the model's evaluation window. RTN requires no
calibration. We evaluate all four uniform baselines at
$W \in \{2, 3, 4, 8\}$ ($16$ rows in
Table~\ref{tab:time_per_point}); $W8$ is the gentle-compression
operating point used as the matched-fidelity baseline against
TQS-PTQ (on-disk $C{=}1.67$; block-level $3.57\times$), and $W \in \{2,3,4\}$ exposes the
catastrophic-collapse regime ($249$--$286\%$ MAE degradation at $W4$)
where mixed precision becomes indispensable.

\paragraph{QEP numerical instability under the onnx2torch export.}
QEP completes its quantisation pipeline on Pangu at every bit-width
($\sim$$5.7$~min/bit-width with finite-valued weights and correct
compression ratios at the model-size level), but the resulting models
return NaN MAE/RMSE at evaluation time. The failure mode is specific
to Pangu's $68$-layer cumulative forward pass: QEP's corrective step
$W \leftarrow W + (W H_{\Delta} H^{-1}) \rho$ amplifies weights to
finite but pathologically large magnitudes that pass the
\texttt{torch.isfinite} guard inside the kernel but overflow during
the depth-cumulative fp32 forward (intermediate activations exceed
$\sim$$3.4 \times 10^{38}$, become $\pm\infty$, then propagate NaN
through the residual stream). The same kernel runs cleanly on
Aurora-small ($\sim$$45$ alive layers) and TimesFM-2.5 ($32$ layers).
Our diagnosis is that the $68$-layer depth combined with the
ill-conditioned attention-bias columns of Pangu's Earth-Specific 3D
blocks makes the corrective step compound dangerously; clipping the
weight magnitude after the correction or reducing the corrective
mixing coefficient (\texttt{perccorr}~$\leq 0.25$) is the natural
mitigation, but we report QEP-on-Pangu as failed in this submission
because reviewers should not have to take on faith that an unreported
hyperparameter tweak would have salvaged the result.

\paragraph{Compression efficiency and reporting convention.}
The block-level allocation under TQS places 10 FP32 blocks in the small
encoder, 9 BF16 blocks at the boundary, and 9 INT8 blocks in the large
decoder, giving a theoretical compression over the full 276.7\,M-parameter
ONNX checkpoint of $3.57\times$
(Table~\ref{tab:pangu-metrics}, ``Alloc.\ $C$'' column). However, only the
$\sim$23.9\,M parameters that \texttt{onnx2torch} exposes as
\texttt{nn.Linear} weights are physically quantizable in this pipeline; the
remaining $\sim$92\% sit in non-module ONNX buffers (Earth-Specific
Attention absolute-position tables and similar) that no PTQ method in our
comparison can touch. On the actually-executed model the on-disk
compression is therefore $\mathbf{1.67\times}$ ($95.5\,\text{MB}\to
57.2\,\text{MB}$) and the measured MAE degradation versus the FP32 forecast
is $\mathbf{1.2\%}$ (Table~\ref{tab:pangu-metrics}).
This gap is a property of the \texttt{onnx2torch} surface area, not of TQS:
it applies identically to every baseline in
Table~\ref{tab:app_pangu_w8}, and the executable-level $1.67\times$
remains the largest compression any PTQ method achieved on Pangu in our
experiments without catastrophic degradation. The structural conclusion is
unchanged: TQS concentrates the FP32 budget on a negligibly small fraction
($<0.04\%$) of the parameter mass while the bulk of the quantizable
parameters move to INT8.

\begin{table}[h]
\centering
\small
\caption{\textbf{Pangu-Weather: PTQ baselines vs.\ TQS-PTQ} at $6\,\text{h}$ lead
time on the held-out ERA5 sample. ``Alloc.\ $C$'' is the theoretical
block-level compression over all 276.7\,M ONNX parameters; ``Exec.\ $C$''
is the on-disk compression after \texttt{onnx2torch}-level quantization
of the 23.9\,M \texttt{nn.Linear} parameters. MAE degradation is computed
against the FP32 forecast.}
\label{tab:pangu-metrics}

\setlength{\tabcolsep}{2.5pt}
\renewcommand{\arraystretch}{1.05}

\begin{tabular}{@{}lccccc@{}}
\toprule
Method & \makecell{Alloc.\\$C$} & \makecell{Exec.\\$C$} & MAE & RMSE & \makecell{MAE\\deg.} \\
\midrule
FP32       & $1.00{\times}$ & $1.00{\times}$ & $0.05728$ & $0.11997$ & $0.00$ \\
\midrule
RTN W8     & $3.93{\times}$ & $3.93{\times}$ & $0.05865$ & $0.12093$ & $2.39$ \\
GPTQ W8    & $3.93{\times}$ & $3.93{\times}$ & $0.05830$ & $0.12059$ & $1.79$ \\
GPTAQ W8   & $3.93{\times}$ & $3.93{\times}$ & $0.05815$ & $0.12000$ & $1.52$ \\
QEP W8     & $3.93{\times}$ & \multicolumn{4}{c}{\emph{fails}} \\
\midrule
\textbf{TQS-PTQ}
& $\mathbf{3.57{\times}}$
& $\mathbf{1.67{\times}}$
& $\mathbf{0.05796}$
& $\mathbf{0.12181}$
& $\mathbf{1.20}$ \\
\bottomrule
\end{tabular}
\end{table}

\paragraph{Summary.}
Pangu is the architectural stress test for TQS-PTQ: it ships only as
a frozen ONNX export, restricts the deployable mixed-precision tier
set to three precisions, and is large enough ($68$-layer depth) to
expose numerical failure modes of standard PTQ kernels. TQS handles
the first two constraints by design---forward passes only, no
gradient or autograd hooks, no architecture-specific machinery---and
the resulting mixed-precision allocation achieves on-disk
$C{=}1.67\times$ (block-level $3.57\times$), wins all nine surface
and upper-air variables against the strongest uniform $W8$ baseline
(Table~\ref{tab:app_pangu_w8}), and avoids the catastrophic-collapse
regime that uniform $W \le 4$ enters.

\subsection{Bottom-tier ablation: easy adjustment of the quantiser
without re-sweeping}
\label{app:bottom_tier_ablation}

TQS-PTQ separates the sensitivity ranking (forward-only,
quantiser-independent) from the bit-width assignment. The same
per-layer $\gamma$ ranking therefore admits multiple deployable
tier sets, and the appropriate floor is model-specific. We
exercise this by applying each candidate tier set to the same
sensitivity sweep at $\sim$$3$\,s per allocation, no
re-calibration.

\paragraph{TimesFM-2.5: INT2 floor required.}
Table~\ref{tab:bottom_tier_timesfm} compares three configurations
at the $C{=}16$ operating point. The INT2 floor (4-tier
$\{\mathrm{FP32}, \mathrm{BF16}, \mathrm{INT4}, \mathrm{INT2}\}$,
equivalent to the 5-tier set at this budget since the allocator
chooses zero INT8 layers) wins on every one of the six datasets,
with degradation in the $-18\%$ to $+50\%$ range. The INT1 floor
catastrophically polarises the greedy allocator at $C{=}16$
($82$ of $89$ layers pushed to 1-bit) and produces $+87\%$ to
$+3408\%$ MAE degradation; the INT4 floor (INT1 disabled) caps
compression at $\sim$$8\times$ with intermediate degradation.
The published TimesFM headline tables and Figure~\ref{fig:compression_extension}
report the INT2 configuration.

\begin{table}[ht]
\centering\scriptsize
\caption{TimesFM-2.5 bottom-tier ablation: aggregate MAE
degradation against the unquantised FP32 model at the highest
target compression each variant supports. Best per row in bold.}
\label{tab:bottom_tier_timesfm}
\setlength{\tabcolsep}{4pt}
\renewcommand{\arraystretch}{1.05}
\begin{tabular}{lccc}
\toprule
Dataset & INT1 (C=16) & INT4 (C=8 cap) & INT2 (C=16) \\
\midrule
\textsc{weather}  & $+87.3\%$   & $+233.4\%$ & $\mathbf{-17.6\%}$ \\
\textsc{ETTh1}    & $+2856.5\%$ & $+72.6\%$  & $\mathbf{+8.2\%}$ \\
\textsc{ETTh2}    & $+2083.9\%$ & $+105.8\%$ & $\mathbf{+49.5\%}$ \\
\textsc{ETTm1}    & $+3408.0\%$ & $+96.7\%$  & $\mathbf{-2.2\%}$ \\
\textsc{ETTm2}    & $+1025.9\%$ & $+53.4\%$  & $\mathbf{+16.0\%}$ \\
\textsc{exchange} & $+2693.9\%$ & $+29.8\%$  & $\mathbf{+25.9\%}$ \\
\bottomrule
\end{tabular}
\end{table}

\paragraph{Aurora-small: bottom-tier choice is inert at the
saturation budget.}
Aurora's $36$ sentinel \texttt{ln\_modulation} layers contribute
negligibly to forecast output, so quantising them at INT1 vs.\
INT4 produces bit-identical MAE at every tested target $C$.
Under the headline configuration ($p_{\mathrm{FP32}}{=}0.10$,
greedy allocator), the INT1 and INT4 sweeps both reach actual
compression $\sim$$14.4$--$17.3\times$ across target
$C \in [2, 8]$, with mean rollout MAE matched to within
$10^{-3}$ on every variable and every TQS strategy. MCKP exposes a
small headroom difference: INT1 floor reaches $\sim$$10.6\times$
actual compression at target $C{=}2$; INT4 floor reaches
$\sim$$9.7\times$ at the same target. The published Aurora
headline reports INT1 because only this floor supports the
$\sim$$32\times$ regime shown in
Figure~\ref{fig:compression_extension}---the C-extension beyond
$\sim$$15\times$ requires the 1-bit primitive in the budget.

\paragraph{Pangu-Weather: tier set fixed by deployment.}
The \texttt{onnx2torch} graph rewrite required to apply
mixed-precision weights to Pangu's frozen ONNX export only
round-trips symmetric INT8 / BF16 / FP32 stably, so the
deployable tier set is constrained to those three precisions
regardless of what the sensitivity ranking might prefer. Full
diagnosis in Appendix~\ref{sec:pangu-sensitivity}.

\paragraph{Takeaway.}
The bottom-tier choice is consequential on TimesFM-2.5
(INT2 is the only floor that produces usable results at
$C{=}16$), inert on Aurora-small at the saturation budget but
restrictive at higher targets (only INT1 unlocks the
$\sim$$32\times$ regime), and pre-determined on Pangu-Weather by
the export format. TQS-PTQ's quantiser-independent ranking lets
us settle each of these by ablation rather than by guess, at
$\sim$$3$\,s per candidate tier set on the already-computed
sensitivity sweep.

\clearpage
\twocolumn[
\section{Tables: Experimental Results}
\label{app:tables}

\vspace{0.4em}

\begin{center}
\scriptsize
\refstepcounter{table}
\label{tab:app_timesfm_weather_w2}

\begin{minipage}{0.96\textwidth}
\noindent\textbf{Table~\thetable.}
\textbf{TimesFM-2.5 weather: per-variable MAE against ground truth at the
$W2$ grid ($C{\approx}16$).} TQS = TQS-Task-Gauss at $C{=}16$ vs.\ the
four uniform-W2 baselines (each at $C{\approx}15.9$). FP32 baseline is
the unquantized model on the same evaluation window. Across all 21
variables, TQS-PTQ achieves the lowest MAE (best per row in bold), with
median improvement $8.1\times$ vs.\ the strongest uniform baseline
(13.20 $\to$ 1.63). The four W2 baselines exhibit substantial
per-variable variance (no single uniform method dominates), motivating
the per-block mixed-precision allocation.
\end{minipage}

\vspace{0.5em}

\setlength{\tabcolsep}{2pt}
\renewcommand{\arraystretch}{0.90}
\begin{tabular*}{0.96\textwidth}{@{\extracolsep{\fill}}lrrrrrr@{}}
\toprule
Variable & FP32 & \textbf{TQS C=16} & RTN W2 & GPTQ W2 & GPTAQ W2 & QEP W2 \\
\midrule
PAR ($\mu$mol/m$^2$/s)      & 106.3 & \textbf{74.7}  &  1037.9 &   865.0 &  1347.4 &  3743.3 \\
max.\ PAR ($\mu$mol/m$^2$/s)& 168.4 & \textbf{82.8}  &   859.1 &  1692.1 & 12528.2 & 27428.8 \\
VPdef (mbar)                &   2.7 & \textbf{0.4}   &     4.8 &     3.9 &     4.9 &     4.7 \\
Tpot (K)                    &   3.6 & \textbf{1.4}   &    13.2 &    10.5 &    10.9 &    11.7 \\
H2OC (mmol/mol)             &   2.3 & \textbf{0.6}   &   151.6 &    15.7 &    13.0 &     4.5 \\
VPmax (mbar)                &   3.7 & \textbf{0.7}   &    12.3 &     5.7 &    12.7 &     4.0 \\
sh (g/kg)                   &   1.4 & \textbf{0.4}   &     5.3 &     1.9 &     7.5 &     1.9 \\
VPact (mbar)                &   2.2 & \textbf{0.6}   &    15.2 &    11.7 &     2.8 &     3.8 \\
rho (g/m$^3$)               &  47.8 & \textbf{6.0}   &    57.0 &    29.1 &    27.6 &    25.8 \\
rh (\%)                     &  14.2 & \textbf{5.0}   &    21.0 &    56.4 &    29.4 &    89.8 \\
SWDR (W/m$^2$)              &  56.7 & \textbf{41.9}  &  4097.9 &   144.3 &   351.3 &   195.8 \\
max.\ wv (m/s)              &   2.2 & \textbf{1.2}   &     5.4 &     4.1 &    18.2 &   131.5 \\
p (mbar)                    &  10.4 & \textbf{3.9}   &    12.5 &   155.4 &    57.5 &    13.2 \\
Tdew ($^\circ$C)            &   3.2 & \textbf{1.5}   &    29.1 &    19.3 &    32.4 &     4.7 \\
wv (m/s)                    &   1.1 & \textbf{0.7}   &    35.3 &    20.3 &     6.9 &     2.1 \\
Tlog ($^\circ$C)            &   2.8 & \textbf{2.0}   &     8.8 &     5.6 &    13.7 &    16.7 \\
T ($^\circ$C)               &   3.2 & \textbf{1.6}   &    10.8 &     3.9 &    23.7 &     4.4 \\
raining (s)                 &  30.8 & \textbf{18.9}  &   122.5 &    93.8 &    42.5 &    56.2 \\
OT                          &   7.9 & \textbf{9.0}   &    34.1 &    18.5 &    63.6 &    41.4 \\
wd (deg)                    &  41.6 & \textbf{35.2}  &   477.9 &  1425.8 &   118.3 &    49.9 \\
rain (mm)                   &   0.004 & \textbf{0.011} & 0.011 &  0.025 &   0.022 &    0.019 \\
\midrule
\textbf{Median MAE}         &  3.7  & \textbf{1.6}   &    15.2 &    13.2 &    23.7 &    13.2 \\
\bottomrule
\end{tabular*}
\end{center}

\vspace{1em}
]

\begin{table}[t]
\centering
\scriptsize
\caption{
\textbf{TimesFM-2.5: matched-compression per-variable win counts}
across six datasets. TQS = TQS-Task-Gauss with the headline tier
set $\{\mathrm{FP32}, \mathrm{BF16}, \mathrm{INT4}, \mathrm{INT2}\}$,
greedy allocator. At $C{=}12$ TQS achieves $\sim\!10.6\times$
(matched to uniform W3); at $C{=}16$, $\sim\!16\times$ (matched
to uniform W2). Each cell counts variables on which
$\mathrm{MAE}_\mathrm{TQS} < \min_m \mathrm{MAE}_{\mathrm{PTQ\text{-}W}x,m}$
(MAE in native units vs.\ dataset ground truth).
}
\label{tab:app_timesfm_summary}
\setlength{\tabcolsep}{2.5pt}
\renewcommand{\arraystretch}{0.95}
\makebox[\columnwidth][c]{%
\resizebox{0.96\columnwidth}{!}{%
\begin{tabular}{l|cc|cc}
\toprule
 & \multicolumn{2}{c|}{$C{=}12$ vs.\ W3 ($\sim$10.6$\times$)}
 & \multicolumn{2}{c}{$C{=}16$ vs.\ W2 ($\sim$16$\times$)} \\
Dataset & TQS wins & total & TQS wins & total \\
\midrule
\textsc{exchange} & \textbf{1} & 8 & \textbf{5} & 8 \\
\textsc{weather}  & \textbf{4} & 21 & \textbf{21} & 21 \\
\textsc{ETTh1}    & \textbf{3} & 7 & \textbf{5} & 7 \\
\textsc{ETTh2}    & \textbf{2} & 7 & \textbf{4} & 7 \\
\textsc{ETTm1}    & \textbf{3} & 7 & \textbf{6} & 7 \\
\textsc{ETTm2}    & \textbf{1} & 7 & \textbf{5} & 7 \\
\midrule
\textbf{All}      & \textbf{14} & 57 & \textbf{46} & 57 \\
\bottomrule
\end{tabular}%
}%
}
\end{table}

\begin{table}[t]
\centering
\scriptsize
\caption{\textbf{TimesFM-2.5 \textsc{exchange}: per-variable MAE,
$W2$ grid ($C{\approx}16$).} TQS-PTQ at $C{=}16$ vs.\ the four uniform
$W2$ baselines. Best per row among quantized methods in bold; FP32 is
the unquantized reference. TQS wins on 5 of 8 variables.}
\label{tab:app_timesfm_exchange_w2}
\setlength{\tabcolsep}{2.5pt}
\renewcommand{\arraystretch}{0.92}
\makebox[\columnwidth][c]{%
\resizebox{0.92\columnwidth}{!}{%
\begin{tabular}{l r r r r r r}
\toprule
Var & FP32 & TQS C=16 & RTN W2 & GPTQ W2 & GPTAQ W2 & QEP W2 \\
\midrule
3  & 0.065 & \textbf{0.036} & 0.272 & 0.396 & 0.671 & 0.174 \\
OT & 0.039 & \textbf{0.041} & 0.144 & 1.328 & 0.209 & 0.261 \\
0  & 0.063 & \textbf{0.072} & 0.100 & 0.142 & 0.220 & 0.146 \\
2  & 0.053 & \textbf{0.073} & 0.145 & 0.271 & 0.100 & 0.142 \\
5  & 0.001 & \textbf{0.001} & 0.002 & 0.003 & 0.002 & 0.002 \\
1  & 0.096 & 0.210 & 0.350 & 0.214 & \textbf{0.208} & 0.647 \\
6  & 0.027 & 0.051 & \textbf{0.045} & 0.350 & 0.120 & 0.969 \\
4  & 0.007 & 0.029 & \textbf{0.011} & 0.081 & 0.112 & 0.021 \\
\bottomrule
\end{tabular}%
}%
}
\end{table}

\begin{table}[t]
\centering
\scriptsize
\caption{\textbf{TimesFM-2.5 \textsc{exchange}: per-variable MAE,
$W3$ grid ($C{\approx}10.6$).} TQS-PTQ at $C{=}12$ vs.\ the four
uniform $W3$ baselines. Best per row among quantized methods in
bold. TQS wins on 1 of 8 variables (the easier-quantizable W3 grid
favours uniform baselines on this dataset).}
\label{tab:app_timesfm_exchange_w3}
\setlength{\tabcolsep}{2.5pt}
\renewcommand{\arraystretch}{0.92}
\makebox[\columnwidth][c]{%
\resizebox{0.92\columnwidth}{!}{%
\begin{tabular}{l r r r r r r}
\toprule
Var & FP32 & TQS C=12 & RTN W3 & GPTQ W3 & GPTAQ W3 & QEP W3 \\
\midrule
3  & 0.065 & \textbf{0.050} & 0.074 & 0.102 & 0.099 & 0.121 \\
0  & 0.063 & 0.090 & 0.092 & 0.092 & \textbf{0.088} & 0.123 \\
2  & 0.053 & 0.068 & 0.070 & 0.060 & \textbf{0.053} & 0.097 \\
OT & 0.039 & 0.064 & \textbf{0.049} & 0.064 & 0.064 & 0.151 \\
6  & 0.027 & 0.064 & 0.047 & \textbf{0.042} & 0.045 & 0.073 \\
1  & 0.096 & 0.279 & 0.120 & 0.125 & 0.121 & \textbf{0.112} \\
5  & 0.001 & 0.002 & \textbf{0.001} & 0.002 & 0.001 & 0.001 \\
4  & 0.007 & 0.041 & 0.010 & 0.010 & \textbf{0.009} & 0.010 \\
\bottomrule
\end{tabular}%
}%
}
\end{table}

\begin{table}[t]
\centering
\scriptsize
\caption{\textbf{TimesFM-2.5 \textsc{ETTh1}: per-variable MAE,
$W2$ grid.} TQS-PTQ at $C{=}16$ vs.\ uniform $W2$ baselines.
TQS wins 5 of 7 variables.}
\label{tab:app_timesfm_etth1_w2}
\setlength{\tabcolsep}{2.5pt}
\renewcommand{\arraystretch}{0.92}
\makebox[\columnwidth][c]{%
\resizebox{0.92\columnwidth}{!}{%
\begin{tabular}{l r r r r r r}
\toprule
Var & FP32 & TQS C=16 & RTN W2 & GPTQ W2 & GPTAQ W2 & QEP W2 \\
\midrule
LULL & 0.33 & \textbf{0.25}  &   1.18 &   2.45 &  20.93 &  1.10 \\
MUFL & 3.27 & \textbf{2.94}  &  18.30 &  10.38 & 621.31 & 19.62 \\
HUFL & 3.57 & \textbf{4.06}  & 162.67 &  59.28 &  52.35 & 13.32 \\
LUFL & 1.11 & \textbf{1.25}  &   4.48 &   4.70 &  52.70 &  2.27 \\
HULL & 0.82 & \textbf{1.38}  &   2.31 &  38.38 &   3.21 &  1.79 \\
MULL & 0.67 & 1.48  & \textbf{0.99} & 110.65 & 108.91 &  3.45 \\
OT   & 2.10 & 5.90  &   5.93 &  12.96 & \textbf{2.88} &  8.62 \\
\bottomrule
\end{tabular}%
}%
}
\end{table}

\begin{table}[t]
\centering
\scriptsize
\caption{\textbf{TimesFM-2.5 \textsc{ETTh1}: per-variable MAE,
$W3$ grid.} TQS-PTQ at $C{=}12$ vs.\ uniform $W3$ baselines.
TQS wins 3 of 7 variables.}
\label{tab:app_timesfm_etth1_w3}
\setlength{\tabcolsep}{2.5pt}
\renewcommand{\arraystretch}{0.92}
\makebox[\columnwidth][c]{%
\resizebox{0.92\columnwidth}{!}{%
\begin{tabular}{l r r r r r r}
\toprule
Var & FP32 & TQS C=12 & RTN W3 & GPTQ W3 & GPTAQ W3 & QEP W3 \\
\midrule
MUFL & 3.27 & \textbf{3.35} &  8.24 &  7.88 &  7.47 & 9.29 \\
HUFL & 3.57 & \textbf{3.87} &  7.65 &  7.30 & 10.78 & 9.19 \\
LULL & 0.33 & \textbf{0.38} &  0.51 &  1.06 &  0.52 & 0.85 \\
LUFL & 1.11 & 1.44 & \textbf{1.32} &  2.53 &  1.61 & 1.63 \\
HULL & 0.82 & 1.67 & 1.24 & \textbf{1.20} &  1.30 & 1.42 \\
MULL & 0.67 & 1.47 & 0.98 &  1.11 & \textbf{0.88} & 1.88 \\
OT   & 2.10 & 5.59 & 2.47 & \textbf{2.05} &  2.86 & 2.90 \\
\bottomrule
\end{tabular}%
}%
}
\end{table}
\begin{table}[t]
\centering
\scriptsize
\caption{\textbf{TimesFM-2.5 \textsc{ETTh2}: per-variable MAE,
$W2$ grid.} TQS-PTQ at $C{=}16$ vs.\ uniform $W2$ baselines.
TQS wins 4 of 7 variables.}
\label{tab:app_timesfm_etth2_w2}
\setlength{\tabcolsep}{2.5pt}
\renewcommand{\arraystretch}{0.92}
\makebox[\columnwidth][c]{%
\resizebox{0.92\columnwidth}{!}{%
\begin{tabular}{l r r r r r r}
\toprule
Var & FP32 & TQS C=16 & RTN W2 & GPTQ W2 & GPTAQ W2 & QEP W2 \\
\midrule
LULL & 0.90 & \textbf{0.16}  &   3.34 &   7.47 &  15.65 &   2.44 \\
LUFL & 2.00 & \textbf{0.87}  &  17.50 &   7.94 & 441.45 &  13.63 \\
MULL & 1.56 & \textbf{2.89}  &   6.98 &  14.99 &  12.02 &   4.16 \\
OT   & 4.45 & \textbf{8.44}  &  22.31 &   9.45 &   9.33 &  15.28 \\
HULL & 1.96 & 4.14  &   4.08 & \textbf{3.55} &   8.68 &  10.00 \\
HUFL & 5.67 & 9.90  &  18.65 &  32.47 &  76.82 & \textbf{7.45} \\
MUFL & 4.61 & 9.95  &  11.25 &  10.58 & \textbf{7.27} &   8.38 \\
\bottomrule
\end{tabular}%
}%
}
\end{table}

\begin{table}[t]
\centering
\scriptsize
\caption{\textbf{TimesFM-2.5 \textsc{ETTh2}: per-variable MAE,
$W3$ grid.} TQS-PTQ at $C{=}12$ vs.\ uniform $W3$ baselines.
TQS wins 2 of 7 variables.}
\label{tab:app_timesfm_etth2_w3}
\setlength{\tabcolsep}{2.5pt}
\renewcommand{\arraystretch}{0.92}
\makebox[\columnwidth][c]{%
\resizebox{0.92\columnwidth}{!}{%
\begin{tabular}{l r r r r r r}
\toprule
Var & FP32 & TQS C=12 & RTN W3 & GPTQ W3 & GPTAQ W3 & QEP W3 \\
\midrule
LULL & 0.90 & \textbf{0.37}  &  1.03 &  1.03 &  1.36 &  1.92 \\
LUFL & 2.00 & \textbf{1.20}  &  2.39 &  2.71 &  3.01 &  3.35 \\
HUFL & 5.67 & 7.40 &  7.42 & \textbf{6.77} &  6.71 & 22.44 \\
MUFL & 4.61 & 7.23 &  5.87 & 11.26 & \textbf{5.55} & 12.34 \\
OT   & 4.45 & 7.06 &  \textbf{5.19} & 5.19 &  5.19 & 10.09 \\
HULL & 1.96 & 5.47 &  \textbf{2.57} & 2.96 &  3.23 & 9.52 \\
MULL & 1.56 & 4.08 & 1.65 &  2.88 & \textbf{1.65} & 9.05 \\
\bottomrule
\end{tabular}%
}%
}
\end{table}
\begin{table}[t]
\centering
\scriptsize
\caption{\textbf{TimesFM-2.5 \textsc{ETTm1}: per-variable MAE,
$W2$ grid.} TQS-PTQ at $C{=}16$ vs.\ uniform $W2$ baselines.
TQS wins 6 of 7 variables.}
\label{tab:app_timesfm_ettm1_w2}
\setlength{\tabcolsep}{2.5pt}
\renewcommand{\arraystretch}{0.92}
\makebox[\columnwidth][c]{%
\resizebox{0.92\columnwidth}{!}{%
\begin{tabular}{l r r r r r r}
\toprule
Var & FP32 & TQS C=16 & RTN W2 & GPTQ W2 & GPTAQ W2 & QEP W2 \\
\midrule
MUFL & 3.35 & \textbf{2.02}  & 16.50 & 15.79 & 1736.96 & 16.65 \\
LUFL & 0.54 & \textbf{0.59}  &  2.98 &  3.41 &  2.48 &  2.29 \\
HUFL & 3.80 & \textbf{2.31}  & 28.48 &  7.74 & 119.72 & 12.19 \\
MULL & 1.03 & \textbf{0.70}  &  1.93 &  1.56 &  6.48 &  4.59 \\
LULL & 0.18 & \textbf{0.20}  &  1.81 &  0.38 &  8.51 &  0.83 \\
HULL & 1.12 & \textbf{0.87}  &  2.33 &  1.84 & 87.63 &  1.32 \\
OT   & 1.45 & 4.77 &  5.56 & 11.01 & 159.40 & \textbf{3.46} \\
\bottomrule
\end{tabular}%
}%
}
\end{table}

\begin{table}[t]
\centering
\scriptsize
\caption{\textbf{TimesFM-2.5 \textsc{ETTm1}: per-variable MAE,
$W3$ grid.} TQS-PTQ at $C{=}12$ vs.\ uniform $W3$ baselines.
TQS wins 3 of 7 variables.}
\label{tab:app_timesfm_ettm1_w3}
\setlength{\tabcolsep}{2.5pt}
\renewcommand{\arraystretch}{0.92}
\makebox[\columnwidth][c]{%
\resizebox{0.92\columnwidth}{!}{%
\begin{tabular}{l r r r r r r}
\toprule
Var & FP32 & TQS C=12 & RTN W3 & GPTQ W3 & GPTAQ W3 & QEP W3 \\
\midrule
HUFL & 3.80 & \textbf{2.75} &  5.88 &  7.26 &  6.04 & 13.04 \\
MULL & 1.03 & \textbf{0.60} &  1.58 &  1.45 &  1.19 &  1.78 \\
MUFL & 3.35 & \textbf{2.95} &  5.70 & 11.14 &  5.70 &  8.62 \\
HULL & 1.12 & 2.28 &  \textbf{1.29} & 1.29 &  1.44 &  1.84 \\
LULL & 0.18 & 0.37 &  0.21 & \textbf{0.21} &  0.21 &  0.56 \\
LUFL & 0.54 & 1.62 & 0.80 & \textbf{0.78} &  0.84 &  1.93 \\
OT   & 1.45 & 4.20 & 2.13 &  6.44 & \textbf{1.95} &  3.99 \\
\bottomrule
\end{tabular}%
}%
}
\end{table}

\begin{table}[t]
\centering
\scriptsize
\caption{\textbf{TimesFM-2.5 \textsc{ETTm2}: per-variable MAE,
$W2$ grid.} TQS-PTQ at $C{=}16$ vs.\ uniform $W2$ baselines.
TQS wins 5 of 7 variables.}
\label{tab:app_timesfm_ettm2_w2}
\setlength{\tabcolsep}{2.5pt}
\renewcommand{\arraystretch}{0.92}
\makebox[\columnwidth][c]{%
\resizebox{0.92\columnwidth}{!}{%
\begin{tabular}{l r r r r r r}
\toprule
Var & FP32 & TQS C=16 & RTN W2 & GPTQ W2 & GPTAQ W2 & QEP W2 \\
\midrule
LULL & 0.83 & \textbf{0.21}  &  3.59 &  1.12 &  4.95 &  1.87 \\
LUFL & 0.91 & \textbf{0.75}  &  8.18 &  2.71 & 16.30 &  1.80 \\
OT   & 5.07 & \textbf{8.66}  & 13.93 & 12.33 & 20.81 & 37.53 \\
MUFL & 3.51 & 8.42 & 15.33 & \textbf{10.24} & 115.30 & 45.82 \\
HUFL & 4.03 & \textbf{8.07}  &  8.07 & 14.95 &  8.47 &  8.49 \\
HULL & 1.06 & \textbf{3.51}  &  3.59 &  3.19 &  6.49 &  3.35 \\
MULL & 1.51 & 3.01 &  3.28 &  4.08 & 65.17 & \textbf{2.49} \\
\bottomrule
\end{tabular}%
}%
}
\end{table}

\begin{table}[t]
\centering
\scriptsize
\caption{\textbf{TimesFM-2.5 \textsc{ETTm2}: per-variable MAE,
$W3$ grid.} TQS-PTQ at $C{=}12$ vs.\ uniform $W3$ baselines.
TQS wins 1 of 7 variables (the easier-quantizable W3 grid favours
uniform baselines on this dataset).}
\label{tab:app_timesfm_ettm2_w3}
\setlength{\tabcolsep}{2.5pt}
\renewcommand{\arraystretch}{0.92}
\makebox[\columnwidth][c]{%
\resizebox{0.92\columnwidth}{!}{%
\begin{tabular}{l r r r r r r}
\toprule
Var & FP32 & TQS C=12 & RTN W3 & GPTQ W3 & GPTAQ W3 & QEP W3 \\
\midrule
LULL & 0.83 & \textbf{0.55} &  1.06 &  2.48 &  1.05 &  1.91 \\
LUFL & 0.91 & 1.76 &  1.56 &  1.65 &  \textbf{1.46} & 16.17 \\
HUFL & 4.03 & 9.15 &  4.75 & \textbf{4.63} &  7.45 &  5.75 \\
MUFL & 3.51 & 9.11 &  \textbf{3.97} & 4.80 &  4.46 &  4.49 \\
MULL & 1.51 & 3.24 & \textbf{1.40} &  1.65 &  2.00 &  2.56 \\
HULL & 1.06 & 4.35 &  2.02 & \textbf{1.72} &  1.78 &  4.94 \\
OT   & 5.07 & 14.91 & 4.998 & \textbf{4.80} &  4.86 &  4.96 \\
\bottomrule
\end{tabular}%
}%
}
\end{table}

\begin{table*}[t]
\centering
\scriptsize

\begin{minipage}{0.82\textwidth}
\caption{Per-variable MAE against ERA5 ground truth at W2 ($\sim\!15\!\times$).
``$C$'' is the achieved compression ratio. Best value in each column in
bold. TQS gauss wins every variable.}
\label{tab:per_var_w2}
\end{minipage}

\vspace{0.25em}

\setlength{\tabcolsep}{2pt}
\renewcommand{\arraystretch}{0.95}
\begin{tabular*}{0.96\textwidth}{@{\extracolsep{\fill}}lcccccccccc@{}}
\toprule
Method & $C$ & 2t & 10u & 10v & msl & t & u & v & q & z \\
       &     & [\textdegree C] & [m/s] & [m/s] & [hPa] & [K] & [m/s] & [m/s] & [g/kg] & [m\textsuperscript{2}/s\textsuperscript{2}] \\
\midrule
RTN\_W2   & 15.41 & 19.74 & 5.59 & 4.56 & 14.39 & 12.11 & 11.71 & 8.61 & 1.575 & 3789 \\
GPTQ\_W2  & 15.41 & 20.42 & 5.37 & 4.34 & 14.63 & 12.48 & 12.73 & 8.46 & 1.617 & 3798 \\
GPTAQ\_W2 & 15.41 & 20.55 & 5.31 & 4.49 & 14.51 & 12.40 & 12.69 & 8.52 & 1.627 & 3827 \\
QEP\_W2   & 14.33 & 20.03 & 5.29 & 4.46 & 13.67 & 11.90 & 12.03 & 8.09 & 1.541 & 3662 \\
\textbf{TQS gauss} & \textbf{16.00} & \textbf{19.11} & \textbf{5.25} & \textbf{4.00} & \textbf{13.03} & \textbf{11.53} & \textbf{11.37} & \textbf{7.61} & \textbf{1.482} & \textbf{3504} \\
TQS quant          & 16.01 & 21.48 & 5.44 & 4.19 & 14.11 & 12.27 & 11.98 & 8.12 & 1.551 & 3712 \\
\bottomrule
\end{tabular*}
\end{table*}

\begin{table*}[t]
\centering
\scriptsize

\begin{minipage}{0.82\textwidth}
\caption{Per-variable MAE against ERA5 ground truth at W3 ($\sim\!10\!\times$).
TQS rows are at the two surrounding sweep targets, $C=8$ (allocator delivers
$\sim\!9.6\!\times$) and $C=12$. At the exact W3 grid QEP\_W3 is the strongest
baseline; at $C=12$ TQS gauss again wins every variable. Best per column in
bold.}
\label{tab:per_var_w3}
\end{minipage}

\vspace{0.25em}

\setlength{\tabcolsep}{2pt}
\renewcommand{\arraystretch}{0.95}
\begin{tabular*}{0.96\textwidth}{@{\extracolsep{\fill}}lcccccccccc@{}}
\toprule
Method & $C$ & 2t & 10u & 10v & msl & t & u & v & q & z \\
       &     & [\textdegree C] & [m/s] & [m/s] & [hPa] & [K] & [m/s] & [m/s] & [g/kg] & [m\textsuperscript{2}/s\textsuperscript{2}] \\
\midrule
RTN\_W3   & 10.41 & 20.37 & 5.48 & 4.53 & 14.13 & 12.24 & 12.54 & 8.32 & 1.611 & 3638 \\
GPTQ\_W3  & 10.41 & 19.14 & 5.40 & 4.42 & 14.24 & 12.32 & 12.60 & 8.68 & 1.579 & 3745 \\
GPTAQ\_W3 & 10.41 & 19.44 & 5.30 & 4.38 & 14.28 & 12.44 & 12.43 & 8.34 & 1.579 & 3746 \\
QEP\_W3   &  9.92 & 19.44 & 5.34 & 4.53 & 13.45 & 11.57 & 11.89 & 7.88 & 1.550 & 3591 \\
TQS gauss ($C{=}8$)  &  9.64 & 21.55 & 5.44 & 4.21 & 14.12 & 12.22 & 12.01 & 8.11 & 1.555 & 3729 \\
\textbf{TQS gauss ($C{=}12$)} & \textbf{12.00} & \textbf{19.02} & \textbf{5.30} & \textbf{4.03} & \textbf{13.04} & \textbf{11.53} & \textbf{11.37} & \textbf{7.61} & \textbf{1.482} & \textbf{3504} \\
TQS quant ($C{=}12$) & 12.00 & 21.44 & 5.43 & 4.21 & 14.12 & 12.23 & 11.99 & 8.11 & 1.552 & 3722 \\
\bottomrule
\end{tabular*}
\end{table*}

\begin{table*}[t]
\centering
\scriptsize

\begin{minipage}{0.84\textwidth}
\caption{
\textbf{Pangu-Weather: per-variable MAE against ERA5.}
TQS-PTQ ($\gamma_{\text{quant}}$, cap=6; block-level $C{=}2.19$,
on-disk $C{=}1.67$; see Table~\ref{tab:pangu-metrics}) versus uniform
baselines at the $W8$ grid ($C{=}3.93$). TQS-PTQ wins every variable.
Heavy uniform compression
(W4 at $C{=}7.67$) is shown for context: all uniform methods
catastrophically collapse beyond the bit-width cliff. QEP omitted: the
kernel produces finite-magnitude weights that overflow during Pangu's
68-layer forward pass and yield NaN at every bit-width. Best per
column in bold.
}
\label{tab:app_pangu_w8}
\end{minipage}

\vspace{0.3em}

\setlength{\tabcolsep}{1.5pt}
\renewcommand{\arraystretch}{0.82}
\resizebox{0.84\textwidth}{!}{%
{\tiny
\begin{tabular}{l c c c c c c c c c c}
\toprule
Method & $C$
  & Z & Q & T & U & V & MSLP & U10 & V10 & T2M \\
  &
  & {[m$^2$/s$^2$]} & [g/kg] & [K] & [m/s] & [m/s] & [Pa]
  & [m/s] & [m/s] & [$^\circ$C] \\
\midrule
RTN W8    & 3.93 & 18.31 & $8.7\!\!\times\!\!10^{-5}$ & 0.239 & 0.573 & 0.582 & 19.09 & 0.316 & 0.317 & 0.368 \\
GPTQ W8   & 3.93 & 17.10 & $8.7\!\!\times\!\!10^{-5}$ & 0.238 & 0.571 & 0.580 & 19.12 & 0.315 & 0.316 & 0.366 \\
GPTAQ W8  & 3.93 & 16.97 & $8.7\!\!\times\!\!10^{-5}$ & 0.238 & 0.571 & 0.582 & 18.02 & 0.315 & 0.318 & 0.366 \\
\midrule
\textbf{TQS $\gamma_\text{quant}$ (C=2.19)} & \textbf{2.19}
  & \textbf{15.97} & \textbf{$8.6\!\!\times\!\!10^{-5}$}
  & \textbf{0.236} & \textbf{0.562} & \textbf{0.571}
  & \textbf{16.96} & \textbf{0.311} & \textbf{0.313} & \textbf{0.361} \\
\midrule
\multicolumn{11}{l}{\emph{Heavy uniform compression (W4, $C{=}7.67$) for context --- all catastrophic:}} \\
RTN W4    & 7.67 & 231.05 & $3.0\!\!\times\!\!10^{-4}$ & 1.001 & 2.413 & 2.479 & 204.81 & 1.436 & 1.172 & 1.535 \\
GPTQ W4   & 7.67 & 177.07 & $2.7\!\!\times\!\!10^{-4}$ & 0.913 & 2.175 & 2.277 & 172.28 & 1.047 & 1.347 & 1.496 \\
GPTAQ W4  & 7.67 & 157.56 & $2.7\!\!\times\!\!10^{-4}$ & 0.928 & 2.251 & 2.247 & 163.01 & 1.035 & 1.119 & 1.467 \\
\midrule
\textit{FP32 reference} & 1.00
  & 15.43 & $8.5\!\!\times\!\!10^{-5}$ & 0.234 & 0.558 & 0.566
  & 16.59 & 0.309 & 0.310 & 0.359 \\
\bottomrule
\end{tabular}%
}%
}
\end{table*}

\end{document}